\pdfoutput=1

\documentclass[opre,nonblindrev]{informs3}

\OneAndAHalfSpacedXII

\usepackage{natbib}
 \bibpunct[, ]{(}{)}{,}{a}{}{,}%

 \usepackage{wrapfig,lipsum,booktabs}

\TheoremsNumberedThrough     
\ECRepeatTheorems

\EquationsNumberedThrough    

\MANUSCRIPTNO{}

\usepackage{psfrag,ifthen,multirow,mathrsfs}
\usepackage{epsfig,psfrag,subcaption,float,texnansi,color,tikz,ifthen}
\usepackage{algorithm}
\usepackage{algpseudocode}
\usepackage{microtype}
\usepackage{graphicx}
\usepackage{latexsym,url,afterpage,enumerate,lmodern}
\usepackage[normalem]{ulem}
\usepackage{amsmath,amssymb,amsfonts}
\usepackage[margin=10pt,font=small,labelfont=bf]{caption}
\usepackage{booktabs,xcolor}
\usepackage{IEEEtrantools}
\usepackage{bm}
\usepackage{enumitem}
\usepackage{tabularx}
\usepackage{dsfont}
\usepackage{rotating}
\usepackage{lscape}
\usepackage{wrapfig}
\usepackage{mathtools}

\usepackage{hyperref}

\newcommand{\sets}[1]{\ensuremath{\mathcal{#1}}}

\DeclareMathOperator{\sgn}{sgn}

\def\proof#1{\Trivlist\item[\hspace*{1em}\hskip\labelsep{\it #1\enskip }]\ignorespaces}

\newcommand{\brct}{RCT-B}
\newcommand{\rct}{RCT}

\newcommand{\revision}[1]{{\normalfont #1}}

\begin{document}


\RUNAUTHOR{Justin, Aghaei, G\'omez, and Vayanos}

\RUNTITLE{Learning Optimal Classification Trees Robust to Distribution Shifts}

\TITLE{Learning Optimal Classification Trees \\ Robust to Distribution Shifts}

\ARTICLEAUTHORS{%
\AUTHOR{Nathan Justin}
\AFF{Center for Artificial Intelligence in Society, University of Southern California, Los Angeles, CA 90089, USA, \EMAIL{njustin@usc.edu}}
\AUTHOR{Sina Aghaei}
\AFF{Center for Artificial Intelligence in Society, University of Southern California, Los Angeles, CA 90089, USA, \EMAIL{saghaei@usc.edu}}
\AUTHOR{Andr\'es G\'omez }
\AFF{Department of Industrial and Systems Engineering, University of Southern California, Los Angeles, CA 90089, USA, \EMAIL{gomezand@usc.edu}}
\AUTHOR{Phebe Vayanos}
\AFF{Center for Artificial Intelligence in Society, University of Southern California, Los Angeles, CA 90089, USA, \EMAIL{phebe.vayanos@usc.edu}}
} 

\ABSTRACT{
We consider the problem of learning classification trees that are robust to distribution shifts between training and testing/deployment data. This problem arises frequently in high stakes settings such as public health and social work where data is often collected using self-reported surveys which are highly sensitive to e.g., the framing of the questions, the time when and place where the survey is conducted, and the level of comfort the interviewee has in sharing information with the interviewer. We propose a method for learning optimal robust classification trees based on mixed-integer robust optimization technology. In particular, we demonstrate that the problem of learning an optimal robust tree can be cast as a single-stage mixed-integer robust optimization problem with a highly nonlinear and discontinuous objective. We reformulate this problem equivalently as a two-stage linear robust optimization problem for which we devise a tailored solution procedure based on constraint generation. We evaluate the performance of our approach on numerous publicly available datasets, and compare the performance to a regularized, non-robust optimal tree. We show an increase of up to 12.48\% in worst-case accuracy and of up to~4.85\% in average-case accuracy across several datasets and distribution shifts from using our robust solution in comparison to the non-robust one.
}


\maketitle

\section{Introduction}

Machine learning  techniques are increasingly being used to support decision-making in high-stakes domains, with potentially significant societal impacts. In such settings, the ability to gauge a machine learning model's trustworthiness is necessary to obtain stakeholder buy-in for deployment. To this end, simple and interpretable models are often preferred over black-box ones~\citep{Rudin2019StopInstead,rudin2022interpretable}.

Some of the most interpretable machine learning models are classification trees, which are easily visualized and simple to deploy.
A classification tree is a model which takes the form of a binary tree. At each branching node, a test is performed, which asks if a feature exceeds a specified threshold value. Given a data sample, if the answer is positive (resp.\ negative) the sample is directed to the right (resp.\ left) descendent. Thus, each data sample, based on its features, follows a path from the root of the tree to a leaf node. At each leaf, a label is predicted. A data sample is correctly classified if and only if its label matches the label predicted at the leaf where it lands~\citep{Breiman2017ClassificationTrees}. 

Training a decision tree consists in deciding which tests to perform at each branching node and which labels to predict at each leaf node, based on some training dataset. A common criterion to evaluate the performance of a decision tree is accuracy, defined as the percentage of correctly classified datapoints (although other metrics are also often employed). In high-stakes settings (e.g., when training a tree to identify those at risk of suicidal ideation, or the most vulnerable among those experiencing homelessness), training to optimality is often critical to ensure that the best model is learned. As long as the data used to train the decision tree comes from the same distribution as that in deployment, then as the size of the training data grows, a tree trained to optimality in this fashion will likely perform as expected in deployment.

Unfortunately, \emph{distribution shifts}, where the distribution of the training and testing/deployment data are different, are common in real-world applications, falling in two broad categories. The first kind of shift corresponds to a change in the likelihood of each sample from the population and occurs when there is sampling bias in the training and/or testing population~\citep{Shimodaira2000ImprovingFunction, Bickel2007DiscriminativeDistributions}, e.g., when using data from Amazon Mechanical Turk\footnote{https://www.mturk.com} to train a model that is then deployed on the entire population. The second kind of shift, which we focus on in the present paper, corresponds to a change in individual entries of the data, often manifesting as a shift in the covariates. This kind of shift occurs when the data collection mechanism changes between training and deployment, or when the environment changes the distribution of data over time, among other reasons \citep{Quinonero-Candela2009DatasetLearning}. Shifts that change the data entries are increasingly prevalent in modern machine learning applications, as full control of data collection in both training and deployment is rare. For example, in domains where data is collected from self-reported surveys, changing the way a question is phrased, the location where the survey is conducted or even the person collecting the information, may result in a shift in responses.

Classification trees, similarly to most machine learning models, are susceptible to distribution shifts, implying that their performance may deteriorate in deployment compared to what was expected from training \citep{zadrozny2004learning}. Thus, methods accounting for potential shifts during the training phase are needed. In this paper, we propose to learn classification trees that are \emph{robust} to distribution shifts, in the sense that they will achieve the best performance under worst-case/adversarial shifts of the training data, resulting in controlled, reliable performance in deployment. 

In social sciences applications where interpretability, robustness, and optimality are often required, the available data frequently come from surveys and manually recorded observations, manifesting as integer and/or categorical data (e.g., gender and race). We thus focus our methodology of learning robust classification trees on the case of covariate shifts on integer and categorical data, a setting of special importance in high-stakes domains that also poses significant modeling and computational challenges.

\subsection{Problem Statement}
\label{sec:problem_statement}

We now formally define the problem of training a classification tree robust to distribution shifts, which is the focus of this paper.

Let~$\{\mathbf{x}^i,y^i\}_{i \in \mathcal{I}}$ be the training set, where~$\mathcal{I}$ is the index set for the training samples and any categorical features are one-hot encoded in each~$\mathbf{x}^i$. The vector~$\mathbf{x}^i~\in~\mathbb{Z}^{|\mathcal{F}|}$ collects the covariates of datapoint~$i$, where the elements of~$\mathcal{F}$ index the features, and~$y^i \in \sets K$ is a label drawn from a finite set~$\mathcal{K}$. We let~$x^i_f$ denote the value of covariate~$f\in \mathcal{F}$ for sample~$i\in \mathcal{I}$ and, with a slight abuse of notation, denote by~$\mathbf{x}$ the vector concatenation of~$\mathbf{x}^i$ over all samples~$i \in \sets I$. Accordingly, we let~$\mathbf{y} := (y^1,y^2,\ldots,y^{|\sets I|})$.

In the presence of distribution shifts, the training set does not reflect the data at deployment. In most instances, the exact shift in the data at deployment is unknown. Thus, it is necessary to account for the performance on several perturbed training sets that reflect some potential distributions of the testing set. For a sample~$i\in \sets I$, let~$\boldsymbol \xi^i \in \mathbb{Z}^{|\mathcal{F}|}$ represent a possible perturbation of the covariates~$\mathbf{x}^i$, such that~$\{ \mathbf{x}^i + \boldsymbol \xi^i, y^i \}_{i \in \sets I}$ is a potential perturbed dataset. With a slight abuse of notation, let~$\boldsymbol  \xi$ be the vector concatenation of~$\boldsymbol \xi^i$ over all samples~$i \in \sets I$. Since~$\boldsymbol \xi$ is unknown in training, we say that~$\boldsymbol \xi$ lies in a set of all possible perturbations~$\Xi$ -- often termed as an uncertainty set in the robust optimization literature \citep{Ben-Tal2009RobustOptimization}.

Let~$\Pi_d$ be the set of all decision trees with depth at most~$d$, where~$d$ is a nonnegative integer. \revision{In settings where interpretability is desired, t}he depth~$d$ is usually chosen to be small, e.g., less than four or so. Each element~$\pi\in \Pi_d$ is a binary decision tree classifier~$\pi:\mathbb{Z}^{|\sets F|} \rightarrow \sets K$. The problem of training an optimal robust classification tree consists in finding a binary tree that correctly classifies the most samples under worst-case realizations of the perturbed data. Mathematically, it is expressible as
\refstepcounter{equation}\label{eq:setup}
\begin{equation}\tag{$\mathcal{P}_\Xi$}\label{eq:single_setup}
        \max_{\pi \in \Pi_d}\; \min_{\boldsymbol \xi \in \Xi} \;\; \sum_{i \in \sets I} \mathbb{I}\left[ \pi(\mathbf{x}^i + \boldsymbol \xi^i) = y^i \right].
\end{equation}
\revision{The above problem is a \emph{single-stage} robust optimization problem, as all decisions, i.e., the classifier~$\pi$, must be decided \emph{before} the uncertain parameters~$\boldsymbol{\xi}$ are observed.}
Note that since the features are integer-valued,~$\Xi$ is a discrete set\revision{, a setting that is not well studied in the robust optimization literature (see section~\ref{sec:discrete_review})}. Also, the summation of indicator functions in~\eqref{eq:single_setup} is nonlinear and discontinuous. These characteristics make problem~\eqref{eq:single_setup} difficult to solve. In this paper, we work to solve problem~\eqref{eq:single_setup} and address these difficulties.

\subsection{Background and Related Work}

Our paper relates closely to several literature streams, which we review in the following.

\subsubsection{MIO-Based Decision Trees}
Traditionally, classification trees are built using heuristic approaches since the problem of building optimal classification trees is~$\mathcal{NP}$-hard \citep{quinlan1986induction, quinlan2014c4, kass1980exploratory, Breiman2017ClassificationTrees}. Recently, motivated by the fact that heuristic approaches often return suboptimal solutions, mathematical optimization techniques such as MIO have been developed for training optimal trees.
The first such method was proposed by \citet{Bertsimas2017OptimalTrees}. To address the long run times required for learning optimal decision trees on large datasets, \citet{Verwer2019LearningFormulation} propose a binary linear programming method whose size scales with the logarithm of the number of training samples. \citet{Aghaei2021StrongTreesb} propose a different MIO method with better relaxation quality to the aforementioned formulations, resulting in improved computational performance. 
\revision{\citet{hu2019optimal} introduce an efficient method for learning optimal decisions trees with sparsity-promoting regularization, giving specialized bounds that can significantly reduce the search space.}
\cite{10.1007/978-3-031-08011-1_7} present a logic-based Benders decomposition method for building optimal trees with multivariate splits. MIO approaches for constructing decision trees allow for several extensions.
Building off of the model by \citet{Bertsimas2017OptimalTrees}, \citet{pmlr-v119-elmachtoub20a} propose an MIO formulation to build optimal decision trees that minimize a ``predict-then-optimize'' loss rather than a classification loss to decrease model complexity while improving decision quality. 
\citet{Misic2020OptimizationEnsembles} examines the setting where the covariates to a given tree ensemble are decision variables, formulating an MIO problem to find the covariates that maximize the ensemble's predicted value. \citet{Aghaei2019LearningDecision-Making} and \citet{jo2023learning} use MIO to create optimal and fair decision trees. \citet{kallus2017recursive} and~\citet{jo2021learning} solve the problem of learning prescriptive trees via MIO, with the latter approach applying also to the case of observational data.

\subsubsection{Machine Learning Robust to Distribution Shifts}\label{sec:ml_robust_review}

Under the setting where there is uncertainty in the parameters of an optimization problem, \emph{robust optimization} has been used to generate solutions immunized to uncertainty by hedging against adversarial realizations of the parameters \citep{Ben-Tal2009RobustOptimization}. Using robust optimization, several researchers have proposed models and algorithms for machine learning that are robust to distribution shifts in the data entries.

\citet{OPT-026} provide a learning framework under distributional perturbations defined by the Wasserstein metric, applying their method to distributionally robust linear regression, semi-supervised learning, and reinforcement learning. More closely related to this paper, there are works on building robust models for a variety of classification tasks. \citet{sinha2020certifying} provide a distributionally robust framework with proven convergence guarantees for smooth loss functions, and apply their framework to neural networks. \citet{shaham2018} create a robust optimization framework for non-parametric models that is trained through a stochastic gradient method, and also apply their method to create artificial neural networks that are robust against adversarial perturbations of the data.
Both robust and distributionally robust optimization approaches have been used to train robust support vector machines \citep{shivaswamy2006second, pant2011support, Bertsimas2019RobustClassification,Kuhn2019WassersteinLearning} and robust logistic regression models \citep{NIPS2015_cc1aa436, Bertsimas2019RobustClassification,Kuhn2019WassersteinLearning}.

\subsubsection{Robust Classification Trees}
\label{sec:rct_review}
Several authors have proposed methods for learning robust classification trees. \citet{chen2019robust} and \citet{vos2021efficient} propose a modification of the standard greedy methods to train decision trees that incorporate robustness. More closely related to our work, \citet{Bertsimas2019RobustClassification} utilize MIO to learn robust classification trees. However, their approach does not solve problem~\eqref{eq:single_setup}, but instead tackles a simpler (conservative) proxy.
We refer to section~\ref{sec:comparisons} for comparisons among these methods from the literature and the approach proposed in this paper.

\subsubsection{Robust Optimization with Discrete Uncertainty Sets}
\label{sec:discrete_review}
The approaches discussed in sections~\ref{sec:ml_robust_review} and~\ref{sec:rct_review} assume that the covariates, and therefore the data perturbations, {are} real-valued. However, real-valued perturbations are usually unrealistic in the high-stakes settings that motivate our work (e.g., in the social sciences and applications using administrative datasets). Additionally, there are very few works in the literature that study problems affected by discrete perturbations. In the case of real-valued perturbations, problem~\eqref{eq:single_setup} is typically handled by dualizing the inner minimization problem. With some exceptions~\citep{bertsimas2003robust, bertsimas2004price, mehmanchi2020robust}, such an approach is {usually} not possible with non-convex {uncertainty} sets (such as ours) due to the lack of strong duality. A possible approach is to approximate the discrete set with a convex relaxation~\citep{atamturk2017maximizing} but such methods {typically} yield poor, conservative solutions~\citep{borrero2021modeling}. In general, solving robust optimization problems with discrete uncertainty sets often requires \revision{iterations between a worst-case analysis} and refinement of a candidate solution. \revision{In particular, the worst-case analysis} typically requires the solution of an expensive mixed-integer optimization problem~\citep{laporte1993integer,mutapcic2009cutting}. 

\subsection{Contribution and Proposed Approach}

We develop the first method for learning optimal robust classification trees.\footnote{
A preliminary version of this paper was presented at the \emph{2022 AAAI Workshop on Adversarial Machine Learning and Beyond}~\citep{justin2022optimal}. The associated paper was peer-reviewed but is `non-archival' and therefore the research presented at the workshop can be published in a journal without conflict.}
A noteworthy characteristic of our method is that, unlike most approaches in the machine learning literature, it allows uncertainty on discrete and/or categorical covariates. We now summarize the key features of our proposed approach.
\begin{enumerate}
    \item We show that the single-stage mixed-integer \emph{nonlinear} robust optimization problem~\eqref{eq:single_setup} can be formulated equivalently as a \emph{two-stage} mixed-integer \emph{linear} robust problem, where the second-stage (recourse) decisions decide on the path followed by each datapoint in the tree and are allowed to adjust to the realization of the data perturbations. We also detail the connections between this two-stage formulation and previous methods for learning robust trees. 
    \item We study the unique setting where the data is discrete, a setting that is of particular relevance in high-stakes applications. To address this problem, we devise a discrete uncertainty set through a cost-and-budget framework. We detail a connection between this proposed uncertainty set and hypothesis testing, giving an informed method for calibrating the uncertainty set.
    \item We present a cutting plane approach that solves the two-stage formulation {to optimality} with the proposed uncertainty set, {allowing the formulation to be implemented in }existing off-the-shelf MIO solvers. 
    \item We evaluate the performance of the formulation on publicly available datasets for several problem instances and show the effectiveness of our proposed method in mitigating the adverse effects of distribution shifts {both on average and in the worst-case}. More specifically, in our computations we observe an increase of up to 12.48\% in worst-case and 4.85\% in average-case accuracy when using a robust tree compared to a non-robust tree in scenarios where the distribution shift is known. The computations also indicate that similar improvements are obtained even if the parameters used to calibrate the uncertainty set are not perfectly known.
\end{enumerate}

The remainder of this paper is organized as follows. Section~\ref{sec:formulation} provides the equivalent reformulation of~\eqref{eq:single_setup} as a two-stage robust linear MIO and describes the model of uncertainty. In section~\ref{sec:solutionmethod}, we describe an algorithm to solve the two-stage formulation to optimality. Section~\ref{sec:stat_connect} provides a way to calibrate the uncertainty set based on hypothesis testing. We also compare the proposed approach to other robust tree formulations in the literature in section~\ref{sec:comparisons}. We then present experimental results in section~\ref{sec:experiments}. \revision{Electronic Companion~\ref{appendix:dro} shows the relationship between problem~\eqref{eq:single_setup} and a Wasserstein distributionally robust version of the problem.} Electronic Companions~\ref{appendix_sec:paths} and~\ref{sec:runtimes_nocut_comparison} contain proofs and empirical results justifying our solution method, respectively. Electronic Companion~\ref{sec:extensions} discusses variants and extensions to our model. Finally, we show additional experimental results in Electronic Companion~\ref{appendix:experiment_bertsimas}. \revision{Note that the approach in this paper has been implemented as part of an open source Python package ODTLearn~\citep{vossler2023odtlearnpackagelearningoptimal}, making it easy to use for researchers and practitioners.}

\section{Robust Tree Formulation}
\label{sec:formulation}
In problem~\eqref{eq:single_setup}, the objective function is highly nonlinear and discontinuous, causing significant difficulties in devising computational solution approaches. We thus propose to reformulate~\eqref{eq:single_setup} equivalently as a two-stage problem with \emph{linear} objective, which will be possible to solve using a conjunction of MIO solvers and delayed constraint generation, see section~\ref{sec:solutionmethod}.

\subsection{Defining a Classification Tree}
\label{sec:tree}
\begin{figure}[t]
\centering
\includegraphics[width=0.65\textwidth]{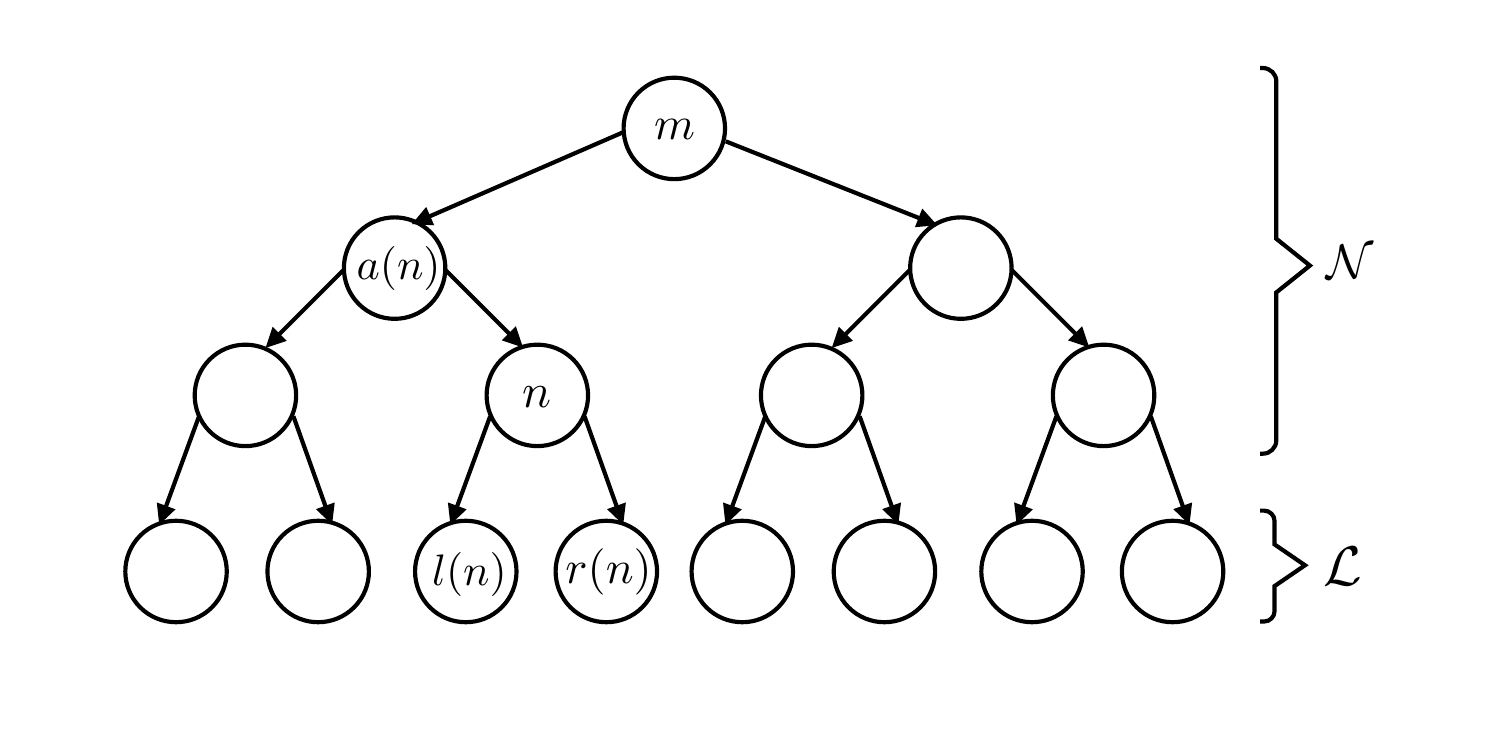}
\caption{Illustration of our notation on a tree of depth~$d = 3$. The nodes in the bottom row are in set~$\sets L$ and all other nodes are in set~$\sets N$. Each node~$n \in \sets N$ has two children:~$l(n)$ (on the left) and~$r(n)$ (on the right), and each node~$n \in \mathcal{N \cup L} \backslash \{1\}$ has a parent~$a(n)$ and ancestors indexed in set~$\sets A(n) = \{m, a(n)\}$.}
\label{img:treestructure}
\end{figure}

To reformulate problem~\eqref{eq:single_setup} as a two-stage problem, we first express~$\Pi_d$ as a set of points satisfying a finite number of linear constraints over a discrete set.
For a tree of maximum depth~$d$, let~$\mathcal{N}$ be the set of internal nodes and~$\mathcal{L}$ be the set of leaf nodes.
There are~$2^{d} - 1$ nodes in~$\mathcal{N}$ and~$2^d$ nodes in~$\mathcal{L}$ and each node is numbered from 1 to~$2^{d+1} - 1$ in a breadth-first search pattern. For any internal node~$n \in \sets N$, let the left and right children of~$n$ be~$l(n)$ and~$r(n)$, respectively. For any node~$n \in \mathcal{N \cup L}$, let~$a(n)$ be the parent of node~$n$ and~$\sets A(n)$ be the set of all ancestor nodes of~$n$. Figure~\ref{img:treestructure} illustrates this notation on a depth 3 classification tree.

Each node in~$\mathcal{L}$ can be a \revision{\textit{prediction node}} where the classification of a sample is made; each node \revision{in}~$\mathcal{N}$ can be a prediction node or a \revision{\textit{branching node}} where a binary test is performed. Any node that is not the root node is \revision{a \textit{pruned node}} (i.e., neither a branching nor prediction node) if any of its ancestors is a prediction node.
At each \revision{branching} node~$n \in \mathcal{N}$, a \revision{binary} test is performed on a chosen feature~$f \in \sets F$ with selected threshold~$\theta$, asking whether the perturbed value of the feature of sample~$i$ is greater than~$\theta$, i.e., if~$x^i_f + \xi^i_f \geq \theta + 1$. If the test passes, then the data sample is directed right to~$r(n)$; otherwise the sample is directed left to~$l(n)$. We let~$c_f$ and~$d_f$ be the lower and upper bounds of values for feature~$f$ within the training data, respectively. The set of possible threshold values we can choose from if branching on feature~$f$ is given by
\[
    \Theta(f) := \left\{\theta \in \mathbb{Z} \ | \ c_f \leq \theta < d_f \right\}.
\]

We now encode elements of~$\Pi_d$ as a discrete set of decision variables for the outer maximization of problem~\eqref{eq:single_setup}. Let~$b_{nf\theta}$ indicate whether node~$n \in \mathcal{N}$ is a branching node and the binary test is on feature~$f \in \mathcal{F}$ with threshold~$\theta \in \Theta(f)$. Binary decision variable~$v_n$ indicates whether node~$n \in \mathcal{N \cup L}$ is a prediction node. Also, we let~$w_{nk}$ be a binary variable that equals one when node~$n \in \mathcal{N \cup L}$ is a prediction node with assignment label~$k \in \mathcal{K}$. Let~$\mathbf{b}$,~$\mathbf{v}$, and~$\mathbf{w}$ collect the~$b_{nf\theta}$,~$v_n$, and~$w_{nk}$ decision variables, respectively. With these variables in hand, we can define a set~$\mathcal{S}$ whose elements have a one-to-one mapping to the trees in~$\Pi_d$:
\begin{subequations}\label{eq:define_S}
    \begin{align} 
        \mathcal{S} \; := \; \{ \  &(\mathbf{b}, \mathbf{v}, \mathbf{w}) : \notag \\
        & \sum_{f \in \mathcal{F}}\sum_{\theta \in \Theta(f)} b_{nf\theta} + v_n + \sum_{m \in \mathcal{A}(n)} v_m = 1 & \forall n \in \mathcal{N}\label{eq:S_internal}\\
        &v_n + \sum_{m \in \mathcal{A}(n)} v_m = 1 & \forall n \in \mathcal{L} \label{eq:S_leaf}\\
        &v_n = \sum_{k \in \mathcal{K}} w_{nk} & \forall n \in \mathcal{N \cup L} \label{eq:S_omega_def}\\
        &b_{nf\theta} \in \{0,1\} & \forall n \in \mathcal{N}, f \in \mathcal{F}, \theta \in \Theta(f)\\
        &v_{n} \in \{0,1\} & \forall n \in \mathcal{N \cup L}\\
        &w_{nk} \in \{0,1\} & \forall n \in \mathcal{N \cup L}, k \in \mathcal{K} \ \}.
    \end{align}
\end{subequations}
Constraints~\eqref{eq:S_internal} state that at each internal node, either a prediction is made at the node, a prediction is made at an ancestor, or a binary test is performed at the node. Constraints~\eqref{eq:S_leaf} affirm that at each leaf node, a prediction is made at either the node or one of its ancestors. Constraints~\eqref{eq:S_omega_def} ensures that exactly one label in~$\mathcal{K}$ is predicted at each prediction node.

\subsection{The Uncertainty Set}
\label{sec:uncertainty_set}

For the time being, we assume that the covariates of all features are integer-valued. We also assume that all uncertain data may shift in either direction by increasing or decreasing in value. We refer to the Electronic Companion~\ref{sec:extensions} for handling categorical features, one-sided shifts that only allow increases (or only decreases) in uncertain data values, and datasets with mixed-feature types and shifts. With these assumptions in place for the inner minimization problem in~\eqref{eq:single_setup}, the uncertainty set~$\Xi$ is defined as
\begin{equation}
    \Xi := \left\{\boldsymbol \xi \in \mathbb{Z}^{|\mathcal{I}| \times |\mathcal{F}|}\ : \ \sum_{i \in \mathcal{I}} \sum_{f \in \mathcal{F}} \gamma^i_f \vert \xi^i_f \vert \leq \epsilon\right\},
    \label{eq:generalUncertainty}
\end{equation}
where~$\gamma^i_f \in \mathbb{R}_+$ is the cost of perturbing~$x^i_f$ by one (in either direction) and~$\epsilon$ is the total allowable budget of uncertainty across data samples. Note that~$\gamma^i_f$ can take on different values for different samples~$i \in \mathcal{I}$ and features~$f \in \mathcal{F}$, which is useful in domains where the likelihood of the covariate shift varies between different groups of samples and/or different features.

As we show in section~\ref{sec:stat_connect}, there exists a connection between~\eqref{eq:generalUncertainty} and hypothesis testing, which allows for calibration of the values of~$\gamma^i_f$ and~$\epsilon$ based on domain knowledge about the possible distribution shifts. \revision{For details on the connection between~\eqref{eq:generalUncertainty} and a Wasserstein-based ambiguity set, we refer to Electronic Companion~\ref{appendix:dro}.
}

\subsection{Counting the Number of Correctly Classified Data Samples}
\label{sec:counting}

We wish to reformulate the expression for the count of correctly classified data samples in~\eqref{eq:single_setup} to create a more computationally tractable formulation. To represent the objective, we use the idea outlined by \citet{Aghaei2021StrongTreesb} that the number of correctly classified samples can be represented as the optimal value of a sum of maximum flow problems, where we adapt this approach to account for perturbations in the data.

\subsubsection{Capacitated Flow Graph for Sample~$i$}

To determine whether a data sample~$i$ is correctly classified for a given tree~$(\mathbf{b}, \mathbf{v}, \mathbf{w}) \in \sets S$ and perturbation~$\boldsymbol \xi \in \Xi$, we define a \emph{capacitated flow graph for sample~$i$} based on~$(\mathbf{b}, \mathbf{v}, \mathbf{w})$, $\mathbf{x}^i$, and $\boldsymbol{\xi}^i$: 

\begin{definition}[Capacitated Flow Graph for Sample~$i$] Let~$\sets{V} := \sets{N \cup L} \cup \{s, t\}$ for~$s$ a source node and~$t$ a sink node. Given tree~$(\mathbf{b}, \mathbf{v}, \mathbf{w}) \in \sets S$, the \emph{capacitated flow graph for sample~$i\in \sets{I}$} is a graph $(\sets{V}, \sets{E})$ such that for \revision{all} $n \in \sets{V} \backslash \{t\}$ and $m \in \sets{V} \backslash \{s\}$, edge $(n,m)$ is in~$\sets E$ and has capacity 1 if and only if \revision{exactly} one of the following is true:
\begin{enumerate}
    \item $n = s$ and $m = 1$;
    \item $n \in \sets{N}$, $m = l(n)$, and there exists an $f \in \sets{F}$ and $\theta \in \Theta(f)$ such that $b_{nf\theta} = 1$ and $x^i_f + \xi^i_f \leq \theta$;
    \item $n \in \sets{N}$, $m = r(n)$, and there exists an $f \in \sets{F}$ and $\theta \in \Theta(f)$ such that $b_{nf\theta} = 1$ and $x^i_f + \xi^i_f \geq \theta + 1$;
    \item  $n \in \sets{N \cup L}$, $m = t$, and $w_{ny^i} = 1$.
\end{enumerate}
Furthermore, edge $(n,m)$ is in $\sets E$ and has capacity 0 if and only if one of the following is true:
\begin{enumerate}
    \item $n \in \sets{N}$, $m = l(n)$, and there exists an $f \in \sets{F}$ and $\theta \in \Theta(f)$ such that $b_{nf\theta} = 1$ and $x^i_f + \xi^i_f \geq \theta + 1$;
    \item $n \in \sets{N}$, $m = r(n)$, and there exists an $f \in \sets{F}$ and $\theta \in \Theta(f)$ such that $b_{nf\theta} = 1$ and $x^i_f + \xi^i_f \leq \theta$;
    \item  $n \in \sets{N \cup L}$, $m = t$, and $w_{ny^i} = 0$.
\end{enumerate}
\end{definition}

\begin{figure}[t]
\centering
\includegraphics[width=0.35\textwidth]{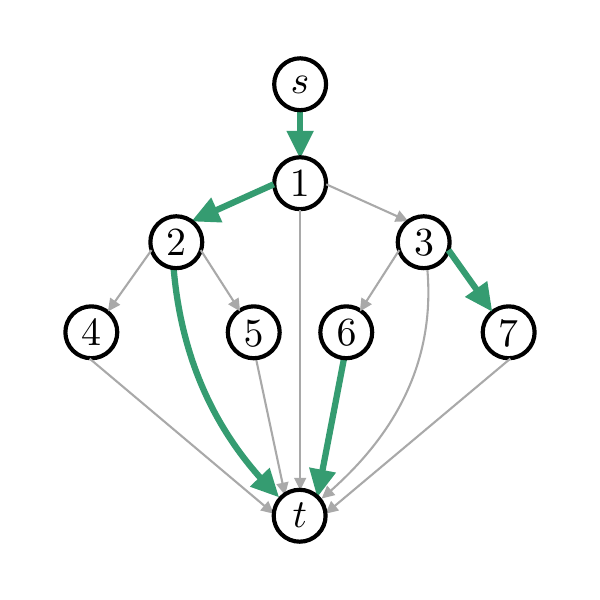}
\caption{
Capacitated flow graph for a given sample~$i$ with a classification tree of depth 2. Nodes~$s$ and~$t$ are the source and sink nodes, respectively,~$\mathcal{N}=\{1,2,3\}$, and~$\mathcal{L}=\{4,5,6,7\}$. The green, bold edges are the edges of capacity 1 in the capacitated flow graph, and the gray edges have capacity 0. Nodes 1 and 3 are branching nodes, where the sample would be routed left and right, respectively. Nodes~2 and~6 assign the correct label, and node 7 assigns an incorrect label. Nodes 4 and 5 are pruned from the tree as they have an ancestor (i.e., 2) that is a prediction node. The maximum flow from~$s$ to~$t$ is 1 in this capacitated flow graph, indicating a correct classification.
}
\label{img:exampletree}
\end{figure}

For every branching node~$n$ with test on threshold~$\theta$ of feature~$f$, the capacitated flow graph has either capacity 0 or 1 for the edge leading to the sink node~$t$ and the edge that is traversed. Specifically, if the test~$x^i_f + \xi^i_f \geq \theta+1$ passes (resp.\ fails), edge~$(n,r(n))$ (resp.~$(n,l(n))$) has capacity 1 and edges~$(n,t)$ and~$(n,l(n))$ (resp.~$(n, r(n))$) have capacity 0. For every prediction node, the edge leading to~$t$ has capacity~1 only if the assigned class of that node is~$y^i$, and all other edges leaving the prediction node have capacity 0. Lastly, as an entry point for a data sample, the edge from the source~$s$ to the root node 1 has capacity~1. This constructed capacitated flow graph for sample~$i$ has a maximum flow from~$s$ to~$t$ of~1 if and only if sample~$i$, perturbed by~$\boldsymbol{\xi}^i$, is correctly classified. Figure~\ref{img:exampletree} illustrates the construction of the capacitated flow graph for sample~$i$.

\subsubsection{Maximum Flow Problem}
We now introduce decision variables that represent the flow of point~$i$ from source~$s$ to~$t$ in the capacitated flow graph. Let the binary decision variable~$z^i_{n,m}$ indicate whether datapoint~$i$ flows down the edge between~$n$ and~$m$ and is correctly classified by the tree for~$m \in \mathcal{N \cup L} \cup \{t\}$ and~$n := a(m) \in \mathcal{N \cup L} \cup \{s\}$ under perturbation~$\boldsymbol \xi$. 
Also let~$\mathbf{z}$ collect all the~$z^i_{a(m),m}$ for all~$i \in \sets I$ and all graph edges~$(a(m),m)$ for~$m \in \mathcal{N \cup L} \cup \{s\}$.
Note that~$z^i_{n,m}$ are the decision variables of a maximum flow problem, where in the capacitated flow graph for sample~$i$,~$z^i_{n,m}$ is~1 if and only if the maximum flow is~1 and the flow traverses arc~$(n,m)$. For fixed~$(\mathbf{b},\mathbf{v},\mathbf{w})$ and~$\boldsymbol \xi$, maximizing the sum of~$z^i_{n,t}$ over all samples~$i \in \sets I$ and all nodes~$n \in \mathcal{N \cup L}$ yields the count of correctly classified samples
\begin{equation} \label{eq:maxflow_obj}
    \max_{\mathbf{z} \in \mathcal{Z}(\mathbf{b}, \mathbf{w}, \boldsymbol \xi)}  \ \sum_{i \in \mathcal{I}}\sum_{n \in \mathcal{N} \cup \mathcal{L}} z_{n,t}^i,
\end{equation}
where the set~$\mathcal{Z}$ defines the maximum flow constraints for each sample's capacitated flow graph:
\begin{subequations}\label{eq:maxflow}
    \begin{align}
        \mathcal{Z}(\mathbf{b}, \mathbf{w}, \boldsymbol \xi) \; := \; \{ \  &\mathbf{z} \in \{0,1\}^{|\mathcal{I}| \times (2^{d+2} - 2)} : \notag \\
        & z_{n, l(n)}^i \leq \sum_{f \in \mathcal{F}}\sum_{\theta \in \Theta(f)} \mathbb{I}\left[x^i_f + \xi^i_f \leq \theta\right]b_{nf\theta} & \forall i \in \mathcal{I}, n \in \mathcal{N}, \label{eq:maxflow_left}\\
        & z_{n, r(n)}^i \leq \sum_{f \in \mathcal{F}}\sum_{\theta \in \Theta(f)} \mathbb{I}\left[x^i_f + \xi^i_f \geq \theta+1\right]b_{nf\theta} & \forall i \in \mathcal{I}, n \in \mathcal{N}, \label{eq:maxflow_right} \\ 
        & z^i_{a(n),n} = z^i_{n, l(n)} + z^i_{n, r(n)} + z^i_{n,t} & \forall i \in \mathcal{I}, n \in \mathcal{N}, \label{eq:maxflow_internal} \\ 
        & z^i_{a(n),n} = z^i_{n,t} & \forall i \in \mathcal{I}, n \in \mathcal{L}, \label{eq:maxflow_leaf} \\ 
        & z^i_{n,t} \leq w_{n, y^i} & \forall i \in \mathcal{I}, n \in \mathcal{N \cup L} \label{eq:maxflow_assignment} \ \}.
    \end{align}
\end{subequations}
Problem~\eqref{eq:maxflow_obj} maximizes the sum of flows over the capacitated flow graphs for all data samples, counting the number of correctly classified samples after perturbation. Constraints~\eqref{eq:maxflow_left} and~\eqref{eq:maxflow_right} are capacity constraints that control the flow of samples based on~$\mathbf{x} + \boldsymbol \xi$ and the tree structure. Constraints~\eqref{eq:maxflow_internal} and~\eqref{eq:maxflow_leaf} are flow conservation constraints. Lastly, constraint~\eqref{eq:maxflow_assignment} blocks any flow to the sink if the node is either not a prediction node or the classification at that node is incorrect.

\subsection{Two-Stage Reformulation}
\label{sec:two_stage}

With the definition of $\mathcal S$ in~\eqref{eq:define_S}, definition of~$\Xi$ in~\eqref{eq:generalUncertainty}, and reformulation of the number of correctly classified datapoints as the optimal value of the maximum flow problem~\eqref{eq:maxflow_obj} in hand, we can rewrite problem~\eqref{eq:single_setup} equivalently as a two-stage \emph{linear} robust optimization problem. In the \emph{first-stage}, the variables~$(\mathbf{b}, \mathbf{v}, \mathbf{w}) \in \sets S$ that encode the tree are selected, corresponding to the outer maximization in problem~\eqref{eq:single_setup}. Once the tree is selected, an adversarial perturbation of the covariates~$\boldsymbol \xi$ from the set~$\Xi$ is chosen, corresponding to the inner minimization in~\eqref{eq:single_setup}. In the \emph{second-stage} problem, given~$(\mathbf{b}, \mathbf{v}, \mathbf{w})$ and $\boldsymbol{\xi}$, the number of correctly classified perturbed samples in the data~$\{ \mathbf{x}^i + \boldsymbol \xi^i, y^i\}_{i \in \sets I}$ is calculated as in~\eqref{eq:maxflow_obj}. This idea leads to the following equivalent reformulation of problem~\eqref{eq:single_setup} for learning optimal robust classification trees:
\begin{equation} \label{eq:primal}
    \max_{(\mathbf{b}, \mathbf{v}, \mathbf{w}) \in \sets S} \; \min_{\boldsymbol \xi \in \Xi} \; \max_{\mathbf{z} \in \mathcal{Z}(\mathbf{b}, \mathbf{w}, \boldsymbol \xi)}  \; \sum_{i \in \mathcal{I}}\sum_{n \in \mathcal{N} \cup \mathcal{L}} z_{n,t}^i.
\end{equation}
Problem~\eqref{eq:primal} is equivalent to problem~\eqref{eq:single_setup}, but unlike formulation~\eqref{eq:single_setup}, it has a linear objective with a linear set of constraints for each stage of the formulation. However,~\eqref{eq:primal} complicates the problem in that it introduces a second stage maximization. Yet in spite of problem~\eqref{eq:primal} being a two-stage problem, this equivalent reformulation of~\eqref{eq:single_setup} can be solved to optimality with the help of MIO solvers, as we show in section~\ref{sec:solutionmethod}.

\revision{
\begin{remark} \label{rem:penalty}
Note that the set~$\mathcal S$ does not involve any constraints that prevent tests that do not improve the tree's accuracy from being performed. To prevent trees that involve such tests from being returned when solving problem~\eqref{eq:primal}, we can add a small penalty term to our objective, ensuring that shallow trees are preferred over deeper trees with the same accuracy. We include such a term in all our experiments (see section~\ref{sec:generate_shifts} for details).
\end{remark}}

\section{Solution Method}
\label{sec:solutionmethod}

We now present a method for solving problem~\eqref{eq:primal} (and therefore problem~\eqref{eq:single_setup}) through a reformulation that can leverage existing off-the-shelf MIO solvers. \revision{There are several exact methods for solving two-stage robust optimization problems. If strong duality holds for the inner maximization problem (as it does in problem~\eqref{eq:primal}), replacing the inner maximization problem with its dual results in an equivalent reformulation with the same optimal objective value and same set of optimal solutions. This dual reformulation is a single-stage robust optimization problem, which can be solved using a delayed constraint generation approach that adds violated constraints during branch-and-cut. We can also use a column-and-constraint generation approach as in~\cite{zeng2013solving}, which iteratively adds both constraints and the inner maximization problem's primal variables to a main problem and resolves the main problem. However, our main problem, represented by the outer maximization of~\eqref{eq:primal}, is difficult to solve as shown in~\citet{Aghaei2021StrongTreesb}. Column-and-constraint generation therefore would necessitate iteratively solving many difficult main problems in succession. Thus, we utilize a tailored delayed constraint generation approach that leverages the structure of the problem, implementable with common off-the-shelf MIO solvers such as Gurobi~\citep{gurobi}.}

\subsection{Reformulating the Two-Stage Problem}
\label{sec:reformulation}

To solve the two-stage optimization problem~\eqref{eq:primal}, we reformulate it equivalently as a single-stage robust MIO. The first part of the reformulation is the dualization of the inner maximization problem~\eqref{eq:maxflow_obj}. Recall that the inner maximization problem is a maximum flow problem; therefore, the dual of the inner maximization problem is a minimum cut problem.

To write the dual, we define the dual variables corresponding to the minimum cut problem. Let~\revision{$q^i_{n, l(n)}$,~$q^i_{n,r(n)}$, and~$q^i_{n,t}$ be the dual variables corresponding to the capacity constraints~\eqref{eq:maxflow_left},~\eqref{eq:maxflow_right}, and~\eqref{eq:maxflow_assignment}, respectively. Each $q^i_{n,m}$ equals 1 if and only if the edge that connects nodes~$n$ and~$m$} in the capacitated flow graph for sample~$i$ is in the minimum cut-set. We let~$\mathbf{q}^i$ be the collection of~$q^i_{n,m}$ over tree edges~$(n,m)$ so that~$\mathbf{q}^i$ represents a cut-set on the capacitated flow graph for data sample~$i$. We also define~$p^i_n$ to be a \revision{dual} variable \revision{corresponding to the flow conservation constraints~\eqref{eq:maxflow_internal} and~\eqref{eq:maxflow_leaf} for $n \in \mathcal{N} \cup \{s\}$ and $n \in \mathcal{L}$, respectively. Each $p^i_n$ equals 1 if and only if node~$n$} is in the source set corresponding to the capacitated flow graph of data sample~$i \in \mathcal{I}$. 

Let~$\mathcal{Q}$ be the set of all possible cut-sets in a classification tree described in section~\ref{sec:tree}, \revision{defined through}
\begin{subequations} \label{eq:mincutsetold}
    \begin{align}
        \mathcal{Q} \; := \; \{ \ & \mathbf{q} \in \{0,1\}^{(2^{d+2} - 2)} \ : \ \exists p_n \in \{0,1\}, \; \mathrlap{\forall n \in \mathcal{N \cup L}\cup\{s\} \; \text{s.t.}} \notag\\
        &  q_{n, l(n)} - p_n + p_{l(n)} \geq 0 & \forall n \in \mathcal{N} \label{eq:mincutset_left}\\
        &  \mathrlap{q_{n, r(n)} - p_n + p_{r(n)} \geq 0 }& \forall n \in \mathcal{N} \label{eq:mincutset_right}\\
        & q_{s,1} + p_1 \geq 1 & \label{eq:mincutset_root}\\
        & -p_n + q_{n, t} \geq 0 &  \forall n \in \mathcal{N \cup L} \label{eq:mincutset_sink} \ \}.
    \end{align}
\end{subequations}
Constraints~\eqref{eq:mincutset_left}-\eqref{eq:mincutset_sink} ensure that if a given node~$n$ is in the source set and one of its children is in the sink set, then the arc connecting the nodes is in the cut-set. Moreover, let~$\mathbf{q} \in \mathcal{Q}^{|\sets I|}$ be a collection of cut-sets~$\mathbf{q}^i \in \mathcal{Q}$ across data samples~$i \in \mathcal{I}$. Then, taking the dual of the inner maximization problem in~\eqref{eq:primal} gives the following single-stage formulation: 

\begin{equation}\label{eq:dual}
    \begin{aligned}
        \max_{(\mathbf{b}, \mathbf{v} ,\mathbf{w})\in \sets S} \ & \min_{\mathbf{q} \in \mathcal{Q}^{|\mathcal I|},\; \boldsymbol \xi \in \Xi} \ \sum_{i \in \mathcal{I}}\sum_{n \in \mathcal{N} \cup \mathcal{L}}q^i_{n, t} w_{n,y^i} + \sum_{i \in \mathcal{I}}q^i_{s, 1} + \sum_{i \in \mathcal{I}}\sum_{n \in \mathcal{N}}\sum_{f \in \mathcal{F}}\sum_{\theta \in \Theta(f)} \mathbb{I}\left[x^i_f + \xi^i_f \leq \theta\right]q^i_{n, l(n)} b_{nf\theta} \\
        & \hspace{25mm} + \sum_{i \in \mathcal{I}}\sum_{n \in \mathcal{N}} \sum_{f \in \mathcal{F}}\sum_{\theta \in \Theta(f)}  \mathbb{I}\left[x^i_f + \xi^i_f \geq \theta+1 \right]q^i_{n, r(n)}b_{nf\theta}.
    \end{aligned}
\end{equation}

Equivalently, we can formulate \eqref{eq:dual} using the hypograph reformulation
\begin{subequations}\label{eq:hypograph_reformulation}
    \begin{align}
        \max_{\mathbf{b}, \mathbf{v}, \mathbf{w},t} \ & t \\
        \text{s.t.} \; \; \; & t \leq \mathrlap{\sum_{i \in \mathcal{I}}\sum_{n \in \mathcal{N} \cup \mathcal{L}}q^i_{n, t} w_{n,y^i} + \sum_{i \in \mathcal{I}}q^i_{s, 1} + \sum_{i \in \mathcal{I}}\sum_{n \in \mathcal{N}}\sum_{f \in \mathcal{F}}\sum_{\theta \in \Theta(f)} \mathbb{I}\left[x^i_f + \xi^i_f \leq \theta\right]q^i_{n, l(n)} b_{nf\theta}} \notag \\
        & \hspace{10mm} + \sum_{i \in \mathcal{I}}\sum_{n \in \mathcal{N}} \sum_{f \in \mathcal{F}}\sum_{\theta \in \Theta(f)}  \mathbb{I}\left[x^i_f + \xi^i_f \geq \theta+1 \right]q^i_{n, r(n)}b_{nf\theta} & \forall \mathbf{q}\in \sets Q^{|\sets I|}, \boldsymbol \xi \in \Xi \label{eq:hypograph_constraints}\\
        & (\mathbf{b}, \mathbf{v},\; \mathbf{w}) \in \sets S \\
        & t \in \mathbb{R},
    \end{align}
\end{subequations}
where decision variable~$t \in \mathbb{R}$ and constraints~\eqref{eq:hypograph_constraints} represent the hypograph of the objective function of~\eqref{eq:dual}. Note that formulation~\eqref{eq:hypograph_reformulation} is a linear MIO formulation for solving problem~\eqref{eq:single_setup}, where the optimal value of~$t$ represents the optimal objective value of~\eqref{eq:single_setup} -- that is, the number of correctly classified datapoints in training under a worst-case realization of the perturbed data. However,~\eqref{eq:hypograph_reformulation} introduces an extremely large number of constraints~\eqref{eq:hypograph_constraints}, one for each combination of cut-set and perturbation in~$\sets Q^{|\sets I|}\times\Xi$, and is impractical to solve directly using MIO solvers.

\subsection{Solving the Single-Stage Reformulation}
\label{sec:cuts}
A common method for solving problems with a large number of constraints is to use a cutting plane approach: solving a simpler relaxation of the original problem as an initial main problem, then iteratively adding constraints as needed. This method typically avoids solving a problem with a prohibitive number of constraints. We propose such a delayed constraint generation approach to solve formulation~\eqref{eq:hypograph_reformulation}.

\subsubsection{The Main Problem}
The approach begins with a main problem that initially relaxes all constraints~\eqref{eq:hypograph_constraints} within formulation~\eqref{eq:hypograph_reformulation}:
\begin{subequations}\label{eq:initmaster}
    \begin{align}
        \max_{\mathbf{b}, \mathbf{v}, \mathbf{w},t} \ & t \\
        \text{s.t.} \; \; \;\; & t \leq |\sets I| & \label{eq:initmaster_t} \\
        & (\mathbf{b}, \mathbf{v}, \mathbf{w}) \in \sets S \\
        & t \in \mathbb{R},
    \end{align}
\end{subequations}
where constraint~\eqref{eq:initmaster_t} bounds~$t$ by its maximum value, i.e., the size of the training set.

Formulation~\eqref{eq:initmaster} can be solved with a linear programming-based branch-and-bound algorithm: a standard approach used to solve MIOs in most solvers. At any integer solution in the branch-and-bound tree, we find and add a violated constraint of the form~\eqref{eq:hypograph_constraints} to~\eqref{eq:initmaster} if one exists. We describe such a process to find a violated constraint in the following section. By adding violated constraints to the main problem along the branch-and-bound process, we converge to the optimal solution of~\eqref{eq:hypograph_reformulation}, which occurs when no violated constraints can be found for the candidate solution and no better integer solutions exist in the branch-and-bound tree. Note that most MIO solvers allow adding constraints as indicated via callbacks.

\subsubsection{The Subproblem} 
We now describe how to check whether a given solution to the relaxed main problem is feasible for \eqref{eq:hypograph_reformulation}, and how to find a violated constraint~\eqref{eq:hypograph_constraints} when infeasible. Given an integer solution of the main problem~$(\mathbf{\bar b},\mathbf{\bar v},\mathbf{\bar w},\bar t)$, we consider the following subproblem:
\begin{equation}
    \begin{aligned}
       \min_{\mathbf{q}\in \sets Q^{|\sets I|},\; \revision{\boldsymbol{\xi}\in \Xi}}\;\Big\{&\sum_{i \in \mathcal{I}}\sum_{n \in \mathcal{N} \cup \mathcal{L}}{q}^i_{n, t} \bar w_{n,y^i} + \sum_{i \in \mathcal{I}}{q}^i_{s, 1} + \sum_{i \in \mathcal{I}}\sum_{n \in \mathcal{N}}\sum_{f \in \mathcal{F}}\sum_{\theta \in \Theta(f)} \mathbb{I}\left[x^i_f + \xi^i_f \leq \theta\right]q^i_{n, l(n)} \bar b_{nf\theta}\\
        & \hspace{5mm} + \sum_{i \in \mathcal{I}}\sum_{n \in \mathcal{N}} \sum_{f \in \mathcal{F}}\sum_{\theta \in \Theta(f)}  \mathbb{I}\left[x^i_f + \xi^i_f \geq \theta+1 \right]q^i_{n, r(n)}\bar b_{nf\theta}\Big\}.
    \end{aligned}
    \label{eq:subproblem}
\end{equation}
Note that the objective function of problem~\eqref{eq:subproblem} is the right-hand side of constraints~\eqref{eq:hypograph_constraints}. Hence, if~$\bar t$ is greater than the optimal value of~\eqref{eq:subproblem}, then the cut~\eqref{eq:hypograph_constraints} where~$(\mathbf{q},\boldsymbol{\xi})$ are fixed to their values in an optimal solution~$(\mathbf{\tilde q},\boldsymbol{\tilde \xi})$ of~\eqref{eq:subproblem} yields a violated constraint that can be added to the main problem.

\revision{Problem~\eqref{eq:subproblem} contains products of terms that jointly optimize over~$\mathbf{q}$ and~$\boldsymbol{\xi}$, presenting its own challenges to solve.
The key observations that enable the effective solution of problem~\eqref{eq:subproblem} given a tree defined by $(\mathbf{\bar b},\mathbf{\bar v},\mathbf{\bar w})$ are that: \emph{1)} Under a given perturbation $\boldsymbol \xi$, each datapoint is either correctly classified or misclassified, contributing 1 or 0 to the objective, respectively; \emph{2)} It is never beneficial to perturb (i.e., to have $\boldsymbol \xi^i \neq \boldsymbol 0$ for) a datapoint $i$ that is already misclassified; \emph{3)} For any given datapoint that is correctly classified, if it is beneficial to misclassify it, we should do so in the minimum cost way; and, \emph{4)} Identifying the minimum cost way of perturbing any given datapoint can be done separately for each datapoint. These observations imply that, for any given tree $(\mathbf{\bar b},\mathbf{\bar v},\mathbf{\bar w})$, we can find the worst-case pair $(\tilde{\mathbf q}, \tilde{\boldsymbol \xi})$ that solves problem~\eqref{eq:subproblem} through a two-step process. First, solve a collection of auxiliary subproblems each of which identifies a pair $(\bar{\mathbf q}^i, \bar{\boldsymbol{\xi}}^i)$ encoding a worst-case path and associated worst-case perturbation that misclassify each datapoint~$i$ at minimum cost (ignoring the perturbation budget $\epsilon$). Second, rank datapoints by minimum cost of misclassification from lowest to highest and perturb as many datapoints as possible according to $\bar{\boldsymbol{\xi}}^i$, starting from the lowest cost ones and without exceeding the budget~$\epsilon$. In the following, we detail and justify this procedure and propose, in particular, an efficient way to identify $(\bar{\mathbf q}^i, \bar{\boldsymbol{\xi}}^i)$ for each datapoint $i$, resulting in a procedure that solves~\eqref{eq:subproblem} in~$\mathcal{O}(|\mathcal{L}||\mathcal{I}| + |\mathcal{I}| \log |\mathcal{I}|)$.
}

\revision{To help formulate the procedure of finding an optimal~$\mathbf{q}$ and~$\boldsymbol{\xi}$ in~\eqref{eq:subproblem}, we now define a \emph{decision path for sample~$i$} and its associated \emph{decision cut-set}~$\mathbf{q}^i$.}
\begin{definition}[Decision Path for Sample~$i$]
    Given a tree~$(\mathbf{\bar b},\mathbf{\bar v},\mathbf{\bar w})$ and sample~$i \in \sets I$, a \emph{decision path} is a sequence of nodes~$(s, n_1, n_2, \ldots, n_k, t) \subseteq \mathcal{N \cup L} \cup \{s,t\}$ for~$k \in \{1, 2, \ldots, d+1\}$ such that, in the capacitated flow graph for sample~$i$,~$n_{j} = a(n_{j+1})$ for all~$j \in \{2, \ldots, k-1\}$, $n_1$ is the root node~$1$, and~$n_k$ is a prediction node (i.e.,~$v_{n_k} = 1$).
\end{definition}
We encode a decision path for sample~$i$ in our subproblem~\eqref{eq:subproblem} through a decision cut-set~$\mathbf{q}^i$, defined as follows.
\begin{definition}[Decision Cut-Set]
    Given a decision path~$(s, n_1, n_2, \ldots, n_k, t) \subseteq \mathcal{N \cup L} \cup \{s,t\}$ and sample~$i$, a \emph{decision cut-set} $\mathbf{q}^i$ is the element of~$\mathcal{Q}$ such that $q^i_{n,m} = 1$ if and only if~$n \in \{n_1, \ldots, n_k\}$ and $m \notin  \{n_2, \ldots, n_k\}$.
\end{definition}
\revision{
\begin{figure}[t]
\centering
\includegraphics[width=0.9\textwidth]{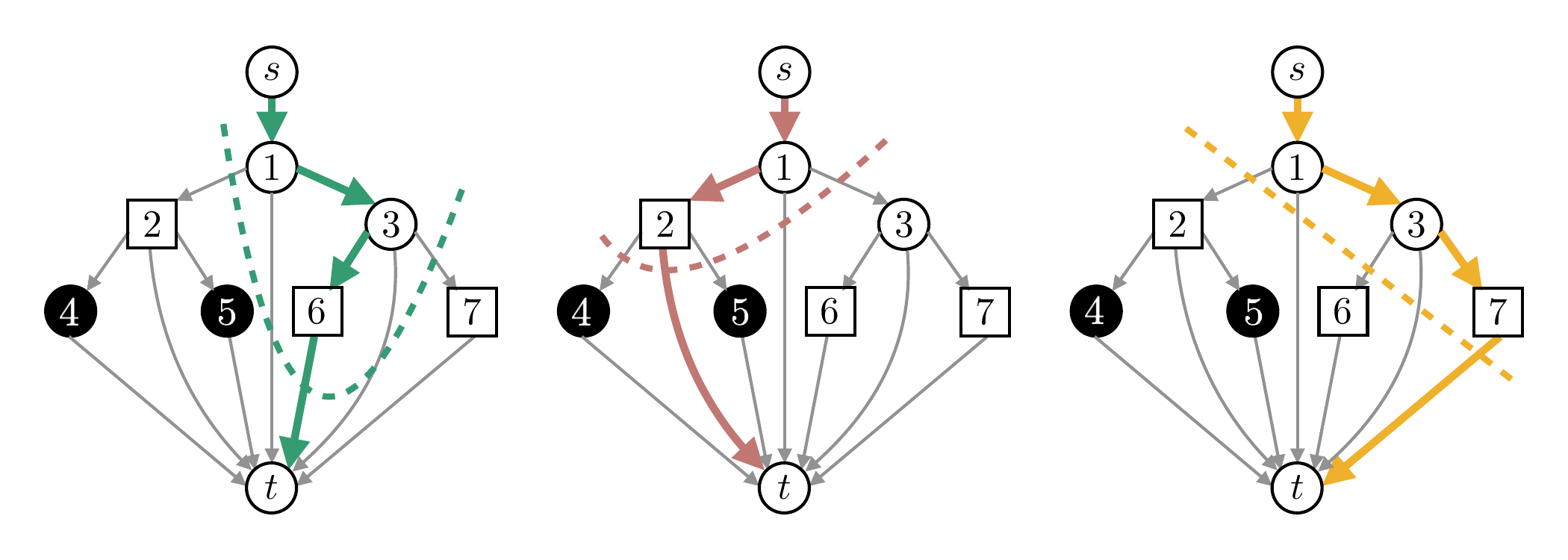}
\caption{\revision{Visual example of all the decision paths and their associated decision cut-sets in the tree described in Figure~\ref{img:exampletree}. Branching nodes are numbered circles (nodes 1 and 3), prediction nodes are numbered squares (nodes 2, 6, and 7), and pruned nodes are filled-in circles (nodes 4 and 5). There are three decision paths in this tree, denoted by the colored, thick edges. For each decision path, its associated decision cut-set $\mathbf{q}^i$ is denoted by the dotted line, where $q^i_{n,m} = 1$ if and only if edge $(n,m)$ crosses the dotted line. For example, the decision path on the left is $(s, 1, 3, 6, t)$ and the associated decision cut-set has $q^i_{1,2} = 1$, $q^i_{1,t} = 1$, $q^i_{3,7} = 1$, $q^i_{3,t} = 1$, $q^i_{6,t} = 1$, and all other $q^i_{n,m} = 0$. Using the example capacitated flow graph in Figure~\ref{img:exampletree}, the left and middle decision cut-sets have value 1, and the right decision cut-set has value 0.}
}
\label{img:decisioncut-ex}
\end{figure}
}
Accordingly, the value of a decision cut-set for sample~$i$ is defined as the sum of edge capacities of the cut-set in the capacitated flow graph for sample~$i$. So, sample~$i$ is misclassified on a decision path if and only if the associated decision cut-set has a value of 0. \revision{We provide examples of decision paths and associated decision cut-sets in Figure~\ref{img:decisioncut-ex}.} 

\revision{To find the minimum cost perturbation that misclassifies a sample~$i$, w}e enumerate over all decision paths that misclassify sample~$i$ and identify the one with the lowest cost perturbation. \revision{For each of these decision paths for sample~$i$}, we identify the minimum cost perturbation that ensures the sample flows through the decision path by solving the following problem\revision{ for~$\boldsymbol{\xi}^i$ given the decision path's associated decision cut-set~$\bar{\mathbf{q}}^i$}:
\begin{equation}\label{eq:cost_perturb}
    \begin{aligned}
       \min_{\boldsymbol{\xi}^i\in \mathbb{Z}^{|\sets F|}}\;&\sum_{f \in \mathcal{F}} \gamma^i_f \vert \xi^i_f\vert\\
        \text{s.t.}\; \; \;& \sum_{n \in \mathcal{N}}\sum_{f\in \sets F}\sum_{\theta\in \Theta(f)} \mathbb{I}\left[x^i_f + \xi^i_f \leq \theta\right]\bar b_{nf\theta}\bar{q}^i_{n, l(n)} + \sum_{n \in \mathcal{N}} \sum_{f\in \sets F}\sum_{\theta\in \Theta(f)}\mathbb{I}\left[x^i_f + \xi^i_f \geq \theta+1 \right]\bar b_{nf\theta}\bar{q}^i_{n, r(n)}=0.
    \end{aligned}
\end{equation}
In this formulation, the objective minimizes the cost of perturbation.
The left-hand side of the constraint corresponds to the objective of~\eqref{eq:subproblem}. The requirement that it equals zero ensures that the choice of~$\boldsymbol{\xi}^i$ follows a decision path associated with decision cut-set~$\bar{\mathbf{q}}^i$ that misclassifies~$i$ (i.e., a path ending at prediction node~$n$ such that~$\bar{w}_{n, y^i} = 0$). For a formal justification of why it is sufficient to solve~\eqref{eq:cost_perturb} at only decision cut-sets~$\bar{\mathbf{q}}^i$, we refer to the Electronic Companion~\ref{appendix_sec:paths}. 

Problem~\eqref{eq:cost_perturb} can be decomposed and solved easily. Indeed, the constraint in~\eqref{eq:cost_perturb} implies that if~$\bar b_{nf\theta}\bar{q}_{n,l(n)}^i>0$, then~$\mathbb{I}\left[x^i_f + \xi^i_f \leq \theta\right]=0$. Similarly, if~$\bar b_{nf\theta}\bar{q}_{n,r(n)}^i>0$, then~$\mathbb{I}\left[x^i_f + \xi^i_f \geq \theta+1\right]=0$. Thus, we can reformulate~\eqref{eq:cost_perturb} as
\begin{equation}\label{eq:cost_perturb2}
\begin{aligned}
       \min_{\boldsymbol{\xi}^i\in \mathbb{Z}^{|\sets F|}}\;&\sum_{f \in \mathcal{F}} \gamma^i_f \vert \xi^i_f\vert\\
        \text{s.t.}\; \; \;& \xi_f^i\geq \theta+1-x_f^i & \revision{\forall f\in \sets F, \; \theta\in \Theta(f) \; : \; \bar b_{nf\theta}\bar{q}_{n,l(n)}^i>0 \text{ for some } n \in \mathcal{N}} \\ 
        & \xi_f^i\leq \theta-x_f^i & \revision{\forall f\in \sets F, \; \theta\in \Theta(f) \; : \; \bar b_{nf\theta}\bar{q}_{n,r(n)}^i>0 \text{ for some } n \in \mathcal{N}.}
        \end{aligned}
\end{equation}

\revision{
The constraints in this problem admit an intuitive interpretation. The first set of constraints states that, if at some branching node $n \in \mathcal N$ we branch on feature $f$ at level $\theta$ ($\bar b_{nf\theta}=1$) and the perturbed datapoint $i$ is routed right ($\bar{q}_{n,l(n)}^i=1$), then it must be that $\xi_f^i + x_f^i \geq \theta+1$. The second set of constraints states that, if at some branching node $n \in \mathcal N$ we branch on feature $f$ at level $\theta$ ($\bar b_{nf\theta}=1$) and the perturbed datapoint $i$ is routed left ($\bar{q}_{n,r(n)}^i=1$), then it must be that $\xi_f^i + x_f^i \leq \theta$. Thus, the path followed by each datapoint (as determined by decision cut-set $\bar{\mathbf{q}}^i$) places constraints on the amount of perturbation for each feature of each datapoint, and the goal of problem~\eqref{eq:cost_perturb2} is to find the smallest cost perturbation that ensures that the perturbed datapoint $i$ is routed through this path. Note that problem~\eqref{eq:cost_perturb2} may be infeasible for certain values of~$\bar{\mathbf{b}}$ and~$\bar{\mathbf{q}}^i$ if there exists a decision path in tree~$(\mathbf{\bar b},\mathbf{\bar v},\mathbf{\bar w})$ that cannot be reached by any value of $\mathbf{x}$. In such cases, we work with the convention that the optimal value of~\eqref{eq:cost_perturb2} is~$\infty$. }

\revision{
Solving problem~\eqref{eq:cost_perturb2} can be done efficiently, noting that the problem decomposes by datapoint and feature. For any fixed sample~$i$ and feature~$f$, determine upper and lower bounds on $\xi_f^i$ by iterating on the nodes of the decision path associated with decision cut-set~$\bar{\mathbf{q}}^i$ and imposing either the first or second constraint on $\xi_f^i$ depending on whether the datapoint is routed left or right; subsequently, determine the value of $\xi_f^i$ from the resulting interval that minimizes $|\xi_f^i|$. Problem~\eqref{eq:cost_perturb2} is infeasible if no~$\xi_f^i$ satisfies the upper and lower bounds.
}

Let~$\psi^i$ be the minimum cost of~\eqref{eq:cost_perturb2} across all possible decision cut-sets \revision{that have value~0 (i.e., decision cut-sets associated with decision paths that misclassify~$i$)}. Once~$\{\psi^i\}_{i \in \sets I}$ has been obtained, we can impose the constraint in~$\Xi$, see~\eqref{eq:generalUncertainty}, which caps uncertainty to a budget~$\epsilon$. Recall that~$\psi^i$ denotes the smallest cost of perturbation to misclassify sample~$i$. Thus, the worst-case set of samples to perturb can be obtained by sorting all non-zero~$\psi^i$ in non-decreasing order, performing the perturbations in this order until the budget~$\epsilon$ is saturated. These perturbations define~$(\mathbf{\tilde q},\boldsymbol{\tilde \xi})$, where~$\mathbf{\tilde q}$ is the collection of decision cut-sets for the decision paths of every sample after perturbation, and~$\boldsymbol{\tilde \xi}$ is the collection of perturbations made (with any unperturbed samples~$i$ after the budget is saturated having perturbation~$\boldsymbol{\tilde{\xi}^i} = \boldsymbol{0}$). The solution~$(\mathbf{\tilde q},\boldsymbol{\tilde \xi})$ then defines the constraint
\begin{equation}\label{eq:violated_constraint}
\begin{split}
    t \; \leq \; &\sum_{i \in \mathcal{I}}\sum_{n \in \mathcal{N} \cup \mathcal{L}}{\tilde q}^i_{n, t} w_{n,y^i} + \sum_{i \in \mathcal{I}}{\tilde q}^i_{s, 1} + \sum_{i \in \mathcal{I}}\sum_{n \in \mathcal{N}}\sum_{f \in \mathcal{F}}\sum_{\theta \in \Theta(f)} \mathbb{I}\left[x^i_f + \tilde{\xi}^i_f \leq \theta\right]\tilde{q}^i_{n, l(n)} b_{nf\theta}\\
        & \hspace{5mm} + \sum_{i \in \mathcal{I}}\sum_{n \in \mathcal{N}} \sum_{f \in \mathcal{F}}\sum_{\theta \in \Theta(f)}  \mathbb{I}\left[x^i_f + \tilde{\xi}^i_f \geq \theta+1 \right]\tilde{q}^i_{n, r(n)} b_{nf\theta}
\end{split}
\end{equation}
which is of the form~\eqref{eq:hypograph_constraints}. The violated constraint~\eqref{eq:violated_constraint} is then added to the main problem. \revision{In summary, solving problem~\eqref{eq:cost_perturb2} requires iterating over all decision paths and data samples, which takes time $\mathcal{O}(|\mathcal{L}||\mathcal{I}|)$, and sorting all $\psi^i$ values requires $\mathcal{O}(|\mathcal{I}| \log |\mathcal{I}|)$ time. This results in an overall runtime of $\mathcal{O}(|\mathcal{L}||\mathcal{I}| + |\mathcal{I}| \log |\mathcal{I}|)$ for the subproblem.} We summarize the procedure for the subproblem in algorithm~\ref{alg:subproblem}.

\begin{algorithm}[tb]
\caption{Subproblem Procedure}
\label{alg:subproblem}
\textbf{Input}: training set indexed by~$\mathcal{I}$ with features~$\mathcal{F}$ and labels~$\mathcal{K}$, uncertainty set parameters~$\boldsymbol \gamma$ and~$\epsilon$, tree~$\pi := (\mathbf{b}, \mathbf{v}, \mathbf{w})$ \\
\textbf{Output}: cut-sets $\tilde{\mathbf{q}}$ and perturbation $\tilde{\boldsymbol{\xi}}$
\begin{algorithmic}[1] 
\For{$i \in \sets I$}
\State \textbf{Initialize: }$\boldsymbol{\xi}^i \leftarrow \boldsymbol 0$ \hfill \Comment{Store optimal perturbation}
\State \textbf{Initialize: }$\psi^i \leftarrow \infty$ \hfill \Comment{Store minimum cost of perturbation}
\State \textbf{Initialize: }$\mathbf{q}^i \leftarrow \boldsymbol 0$ \hfill \Comment{Store optimal cut-set after perturbation}
\For{each decision path $\mathcal{P}$} \hfill \Comment{$\mathcal{P}$ is a path from root node $1$ to a prediction node}
\State $\hat{\mathbf{q}}^i\gets$ decision cut-set associated with $\sets P$
\State $\hat{\psi}^i \leftarrow$ optimal value of $(\ref{eq:cost_perturb}_{\hat{\mathbf{q}}^i, \mathbf{b}})$, $\hat{\boldsymbol{\xi}}^i \leftarrow$ $\argmin\{(\ref{eq:cost_perturb}_{\hat{\mathbf{q}}^i, \mathbf{b}})\}$ \hfill \Comment{Find cost.}
\If{$(\ref{eq:cost_perturb}_{\hat{\mathbf{q}}^i, \mathbf{b}})$ feasible and $\hat{\psi}^i \leq \psi^i$} \hfill \Comment{If infeasible, choose another path}
\State $\boldsymbol{\xi}^i \leftarrow \hat{\boldsymbol{\xi}}^i$, $\psi^i \leftarrow \hat{\psi}^i$, $\mathbf{q}^i \leftarrow \hat{\mathbf{q}}^i$ \hfill \Comment{Store current minimum cost perturbation}
\EndIf
\EndFor
\EndFor
\State \textbf{Initialize:} $C \leftarrow \epsilon$ \hfill \Comment{$C$ is the budget allotted for perturbation}
\State \textbf{Initialize:} $\tilde {\boldsymbol\xi}^i \leftarrow \boldsymbol{0}$ \hfill \Comment{Default perturbation is no perturbation.}
\State \textbf{Initialize:} $\tilde{\mathbf{q}}^i \leftarrow \mathbf{q}^i$ \hfill \Comment{Default cut-set is associated with the nominal path~$\mathbf{q}^i$}
\For{$i \in \sets I$ in non-decreasing order of~$\psi^i$}
    \If{$C - \psi^i >0$}
    \State~$\boldsymbol{\tilde \xi^i} \leftarrow \boldsymbol{\xi}^i$, $\tilde{\mathbf{q}}^i \leftarrow \mathbf{q}^i$ \hfill \Comment{Add perturbations if there is still budget left}
    \State~$C \leftarrow C - \psi^i$ \hfill \Comment{Reduce cost from budget if perturbation of $i$ included.}
    \EndIf
\EndFor
\State \textbf{return} $(\tilde{\mathbf{q}}, \tilde{\boldsymbol{\xi}})$
\end{algorithmic}
\end{algorithm}

\subsection{Improving the Algorithm}
\label{sec:improve_alg}

A direct implementation of the aforementioned delayed constraint generation algorithm may be slow in practice: a prohibitive number of constraints~\eqref{eq:hypograph_constraints} may need to be added before convergence. To improve the performance of the method, we represent the number of correctly classified points, corresponding to the inner mimization in~\eqref{eq:dual}, in an extended formulation. Such representations of nonlinear functions using additional variables often yield stronger relaxations \citep{tawarmalani2005polyhedral,vielma2017extended}, since each linear cut in this lifted space translates to a nonlinear and more powerful cut in the original space. 

In particular, let~$t^i \in \{0,1\}$ be a decision variable that indicates whether sample~$i$ is correctly classified for a given tree~$(\mathbf{b,v,w})$ and worst-case perturbation~$\boldsymbol{\tilde\xi}$ with corresponding minimum cut~$\mathbf{\tilde q}$, and let~$\mathbf{t}$ be the collection of~$t^i$ over~$i \in \mathcal{I}$. 
Since $t$ is the total number of correctly classified points, $t$ is equal to $\sum_{i \in \mathcal{I}} t^i$. So, we reformulate problem~\eqref{eq:hypograph_reformulation} equivalently as
\begin{subequations}\label{eq:additional_cuts_formulation}
    \begin{align}
        \max_{\mathbf{b}, \mathbf{v}, \mathbf{w},\mathbf{t}} \ & \sum_{i \in \mathcal{I}}t^i \\
        \text{s.t.} \;\;\; & \sum_{i \in \mathcal{I}}t^i \leq \mathrlap{\sum_{i \in \mathcal{I}}\sum_{n \in \mathcal{N} \cup \mathcal{L}}q^i_{n, t} w_{n,y^i} + \sum_{i \in \mathcal{I}}q^i_{s, 1} + \sum_{i \in \mathcal{I}}\sum_{n \in \mathcal{N}}\sum_{f \in \mathcal{F}}\sum_{\theta \in \Theta(f)} \mathbb{I}\left[x^i_f + \xi^i_f \leq \theta\right]q^i_{n, l(n)} b_{nf\theta}} \notag \\
        & \hspace{10mm} + \sum_{i \in \mathcal{I}}\sum_{n \in \mathcal{N}} \sum_{f \in \mathcal{F}}\sum_{\theta \in \Theta(f)}  \mathbb{I}\left[x^i_f + \xi^i_f \geq \theta+1 \right]q^i_{n, r(n)}b_{nf\theta} & \forall \mathbf{q}\in \sets Q^{|\sets I|}, \boldsymbol \xi \in \Xi \label{eq:additional_cuts_total}\\
        & (\mathbf{b}, \mathbf{v}, \mathbf{w}) \in \sets S \\
        & t^i \in \mathbb{R} & \forall i \in \mathcal{I}.
    \end{align}
\end{subequations}
Since formulation \eqref{eq:additional_cuts_formulation} is obtained from \eqref{eq:hypograph_reformulation} through the substitution $t = \sum_{i \in \sets I} t^i$, both formulations have identical continuous relaxations. Our approach to further strengthen formulation \eqref{eq:additional_cuts_formulation} relies on one observation: for any datapoint~$i \in \mathcal{I}$ and fixed tree~$\pi \in \Pi_d$ such that $\pi(\mathbf{x}^i) \neq y^i$, there exists an optimal solution to the inner minimization problem in \eqref{eq:single_setup} such that $\pi(\mathbf{x}^i + \boldsymbol \xi^i) \neq y^i$. This is because the inner minimization problem aims to misclassify the most points, so no perturbation is needed on $i$ to misclassify it, i.e., $\boldsymbol{\xi}^i = 0$ is part of the optimal solution.

It follows that we can enforce the condition that any misclassified point~$i$ in the nominal case cannot be correctly classified under a worst-case perturbation via the constraint\revision{
\begin{equation}\label{eq:violated_constraint_single}
    \begin{aligned}
        \revision{t^i} \; \leq \;  &\sum_{n \in \mathcal{N} \cup \mathcal{L}}q^i_{n, t} w_{n,y^i} + q^i_{s, 1} + \sum_{n \in \mathcal{N}}\sum_{f \in \mathcal{F}}\sum_{\theta \in \Theta(f)} \mathbb{I}\left[x^i_f \leq \theta\right]q^i_{n, l(n)} b_{nf\theta}\\
        & \hspace{5mm} + \sum_{n \in \mathcal{N}} \sum_{f \in \mathcal{F}}\sum_{\theta \in \Theta(f)}  \mathbb{I}\left[x^i_f \geq \theta+1 \right]q^i_{n, r(n)}b_{nf\theta} 
        \hspace{10mm} \forall \mathbf{q}^i \in \mathcal{Q}.
    \end{aligned}
\end{equation}
}A derivation of constraints~\eqref{eq:violated_constraint_single} can be found in \citet{Aghaei2021StrongTreesb}. Intuitively, the constraints  are analogous to constraints \eqref{eq:violated_constraint}, but without summing over all datapoints in $\sets I$ and fixing the perturbation $\boldsymbol{\tilde{\xi}}$ to zero. Moreover, separation of inequalities \eqref{eq:violated_constraint_single} is similar to the separation of \eqref{eq:violated_constraint}. \revision{Adding constraints~\eqref{eq:violated_constraint_single} for each decision cut-set~$\tilde{\mathbf{q}}^i$ associated with the decision path of each sample~$i$ without perturbation} improves the solving times by an order of magnitude, which we detail further in Electronic Companion~\ref{sec:runtimes_nocut_comparison}.

\section{Calibration of Uncertainty Set Parameters}
\label{sec:stat_connect}

In this section, we propose a method for calibrating the parameters of the uncertainty set~\eqref{eq:generalUncertainty}. For simplicity, we examine the cases where all features are unbounded integers (or where no bound is known). Datasets with bounded integer features, categorical features, and mixed feature types are discussed in the Electronic Companion~\ref{sec:extensions}.

We calibrate the parameters of~\eqref{eq:generalUncertainty}
with an assumed probability of certainty in the nominal value \revision{of} feature $f$ and training sample~$i$, denoted~$\rho^i_f \in (0,1]$. Without any other knowledge of the distribution shift, we follow the principle of maximum entropy~\citep{jaynes1957information}, which chooses the distribution of the perturbations with greatest entropy and thus highest uncertainty subject to our assumption of the probability of certainty~$\rho^i_f$. To this end, we select the geometric distribution with parameter~$\rho^i_f$ as the distribution of the magnitude of perturbation~$|\xi^i_f|$, and select the direction of perturbation~$\sgn(\xi^i_f)$ uniformly. That is,
for~\revision{$\zeta^i_f \in \mathbb{Z}$} the realization of the perturbation of~$x^i_f$, the probability that~$x^i_f$ is perturbed by~$\xi^i_f$ is given by the symmetric geometric distribution
\begin{equation} \label{eq:symmetric_geometric}
    \mathbb{P}(\xi^i_f = \zeta^i_f) := (0.5)^{\mathbb{I}[|\zeta^i_f|>0]}\rho^i_f(1-\rho^i_f)^{|\zeta^i_f|}
\end{equation}
and the magnitude of perturbation follows a geometric distribution such that
\begin{equation} \label{eq:geometric}
    \mathbb{P}(|\xi^i_f| = |\zeta^i_f|) = \rho^i_f(1-\rho^i_f)^{|\zeta^i_f|}.
\end{equation}
We now follow the idea of building uncertainty sets using hypothesis testing as in~\cite{bertsimas2018data}. We set up a likelihood ratio test on the magnitude of the perturbation with threshold~$\lambda^{|\mathcal{I}|}$ for~$\lambda \in (0,1]$, where we add the exponent~$|\mathcal{I}|$ to normalize across different datasets with varying number of training samples. Our null hypothesis is that the magnitude of a given perturbation~$\boldsymbol \xi$ of our dataset comes from the distribution described by~\eqref{eq:geometric}. If this null hypothesis fails to be rejected, then~$\boldsymbol \xi$ lies within our uncertainty set. Hence,~$\boldsymbol \xi$ lies within our uncertainty set if it satisfies the constraint
\begin{equation}
    \frac{\prod_{i \in \mathcal{I}} \prod_{f \in \mathcal{F}}\rho^i_f(1-\rho^i_f)^{|\xi^i_f|}}{\prod_{i \in \mathcal{I}} \prod_{f \in \mathcal{F}} \rho^i_f} \geq \lambda^{|\mathcal{I}|}.
    \label{eq:ratiotest}
\end{equation}
The numerator of the left hand side of~\eqref{eq:ratiotest} is the likelihood under distribution~\eqref{eq:symmetric_geometric} of a perturbation of magnitude~$|\xi^i_f|$, and thus the likelihood under the null hypothesis. The denominator of the left hand side is the likelihood of the most probable realization under~\eqref{eq:geometric} (i.e., the likelihood of no perturbation \revision{where all $\xi^i_f = 0$}). The test~\eqref{eq:ratiotest} can be equivalently represented as
\begin{equation}
    \sum_{i \in \mathcal{I}} \sum_{f \in \mathcal{F}}|\xi^i_f| \log\left(\frac{1}{1-\rho^i_f}\right) \leq -|\mathcal{I}|\log\lambda.
    \label{eq:hyptest}
\end{equation}
Note that~\eqref{eq:hyptest} is of the form of the constraint in uncertainty set~\eqref{eq:generalUncertainty} with parameters~$\gamma^i_f = \log\left(\frac{1}{1-\rho^i_f}\right)$ and~$\epsilon = -|\mathcal{I}|\log\lambda$.

Therefore, a method of tuning the parameters of~\eqref{eq:generalUncertainty} is to use the \revision{probability of certainty~$\rho^i_f$} for each feature of each sample, which can be derived from domain knowledge. The value of~$\lambda$ is used to tune the size of the uncertainty set, and hence the level of robustness. For~$\lambda = 1$, the budget of uncertainty~$\epsilon$ is~0, meaning that no distribution shift occurs and our formulation is equivalent to the formulation from~\cite{Aghaei2021StrongTreesb}. As~$\lambda$ decreases, more and more perturbations of the data are considered, and thus the model is robust to more scenarios. Hence,~$\lambda$ is a parameter that can be tuned to adjust the level of robustness, which can be chosen by methods such as cross validation. \revision{Lastly, we note that other distributions can be used based on domain knowledge of the possible distribution shift, and that the geometric distribution~\eqref{eq:geometric} is a generic choice given limited domain knowledge.}


\section{Comparison to State-of-the-Art Methods}
\label{sec:comparisons}
In this section, we compare the uncertainty sets and notions of robustness used in state-of-the-art models from the literature and in our model for learning robust classification trees.
In particular, we examine the approach of \cite{Bertsimas2019RobustClassification}, which also utilizes an MIO-based model but with a different uncertainty set and concept of robustness. We also compare and contrast to
\cite{chen2019robust} and \cite{vos2021efficient} who also use a different uncertainty set and employ a heuristic approach, but who use a similar notion of robustness to ours.

\subsection{Uncertainty Sets}
\label{sec:compare_uncertainty}
The model of uncertainty~\eqref{eq:generalUncertainty} that we employ differs in several regards from that in previous works on robust decision trees. Indeed, \cite{Bertsimas2019RobustClassification}, \cite{chen2019robust}, and \cite{vos2021efficient} employ a \emph{row-wise uncertainty set} based on an~$p$-norm, of the form
\begin{equation} 
    \Xi_p = \left\{\boldsymbol \xi \in \mathbb{R}^{|\sets I| \times |\sets F|} \ : \ \Vert \boldsymbol \xi^i \Vert_{\revision{p}} \leq \epsilon \quad \forall i \in \sets I \right\},\label{eq:bertsimas_uncertainty}
\end{equation}
where~$\epsilon \geq 0$ is a user selected budget of uncertainty parameter.  \cite{chen2019robust} and \cite{vos2021efficient} model uncertainty with~$\Xi_\infty$ specifically. On the other hand, \cite{Bertsimas2019RobustClassification} uses the uncertainty set~\eqref{eq:bertsimas_uncertainty} with any choice of~$p$-norm. We note that in \cite{Bertsimas2019RobustClassification}, due to the notion of robustness employed, the robust counterpart ends up taking the same form independently of the choice of~$p$, resulting in the same sets of optimal trees independently of the norm used in the uncertainty set.

We now argue that the uncertainty set in~\eqref{eq:bertsimas_uncertainty} does not model distribution shifts in the datasets and applications that motivate us. First, the perturbation~$\boldsymbol \xi$ is not integral, and thus the realization of the covariates may not be realistic if the covariates are integer or categorical. In addition, the set~$\Xi_p$ imposes the strong assumption that the distribution shift is rectangular across samples, resulting in an overly conservative model where the perturbations associated with all datapoints can all simultaneously take on their worst-case values. Lastly, the set~\eqref{eq:bertsimas_uncertainty} assumes the same cost of shift~(represented by $\gamma^i_f$ in our model) across all datapoints~$i \in \sets I$ and features~$f \in \sets F$, implying that the magnitude and direction of distribution shifts is constant for all samples and features.

In contrast, our proposed uncertainty set~\eqref{eq:generalUncertainty} fixes the aforementioned issues of uncertainty set~\eqref{eq:bertsimas_uncertainty} by restricting the perturbations to be integer, having a single budget of uncertainty shared among the data samples, and introducing costs of perturbation that can differ for each feature and sample. Thus, our proposed model of uncertainty is more flexible, being able to capture shifts in integer covariates and, as discussed in section~\ref{sec:extensions}, extending to the cases of one-sided shifts and categorical features.

\subsection{Notions of Robustness}
\label{sec:compare_robustness}
In our problem, the tree structure must be decided before the perturbation of covariates is realized, and only \emph{after} this realization is observed can we decide if a given datapoint is correctly classified or not.  This is similar to the frameworks of \cite{chen2019robust} and \cite{vos2021efficient}, who calculate their objective based on an adversarial perturbation of the data. But unlike our approach of optimizing accuracy over the whole tree, these methods use either information gain or Gini impurity as an objective at each node where a test is performed. These approaches, therefore, cannot guarantee optimal worst-case accuracy.

On the other hand, \cite{Bertsimas2019RobustClassification} does have accuracy as an objective, but uses a different notion of robustness. Their model postulates that the robust trees created must maintain the \emph{same} predictions for each data sample across \textit{all possible perturbations of the data}. Mathematically, the problem solved by \cite{Bertsimas2019RobustClassification} is equivalent to
\refstepcounter{equation}\label{eq:proxy}
\begin{equation}\tag{$\text{\ref{eq:proxy}}_\Xi$}\label{eq:proxy-xi}
\begin{split} 
        \max_{\pi \in \Pi_d}\; \;& \sum_{i \in \sets I} \mathbb{I}\left[ \pi( {\bf{ x}}^i ) =  y^i \right]\\
        \text{s.t.}\;\;\;&\pi(\mathbf{x}^i)=\pi(\mathbf{x}^i + {\boldsymbol{\xi}}^i) \quad \forall i\in \sets I,\; \boldsymbol \xi \in \Xi.
\end{split}
\end{equation}
It follows from the constraints of~\eqref{eq:proxy-xi} that it is equivalent to
\refstepcounter{equation}\label{eq:proxy_reformulate}
\begin{equation}\tag{$
\text{\ref{eq:proxy_reformulate}}_\Xi$}\label{eq:proxy_reformulate-xi}
\begin{split} 
        \max_{\pi \in \Pi_d}\; \; & \min_{\boldsymbol{\xi} \in \Xi}\; \sum_{i \in \sets I} \mathbb{I}\left[ \pi( {\bf{x}}^i +  {\boldsymbol{\xi}}^i) =  y^i \right]\\
        \text{s.t.}\;\;\;&\pi(\mathbf{x}^i)=\pi(\mathbf{x}^i + {\boldsymbol{\xi}}^i) \quad \forall i\in \sets I,\; \boldsymbol \xi \in  \Xi.
\end{split}
\end{equation}
Note that~\eqref{eq:proxy_reformulate-xi} has the same objective function as~\eqref{eq:single_setup}. We formalize the relationship between problems~\eqref{eq:single_setup} and~\eqref{eq:proxy_reformulate-xi} in the following proposition.

\begin{proposition} \label{prop:subset}
Given an uncertainty set~$\Xi$, all optimal solutions of~\eqref{eq:proxy_reformulate-xi} are feasible in~\eqref{eq:single_setup}, but all optimal solutions to~\eqref{eq:single_setup} may be infeasible in~\eqref{eq:proxy_reformulate-xi}.
\end{proposition}
\proof{Proof.}
Note that problem~\eqref{eq:proxy_reformulate-xi} is problem~\eqref{eq:single_setup} with the additional constraints in~\eqref{eq:proxy_reformulate-xi}. Therefore, problem~\eqref{eq:proxy_reformulate-xi} has a feasible region that is a subset of the feasible region of problem~\eqref{eq:single_setup}. Hence, any optimal solution of~\eqref{eq:proxy_reformulate-xi} is a feasible solution of~\eqref{eq:single_setup}.

We now present an example where all optimal solutions to~\eqref{eq:single_setup} are infeasible in~\eqref{eq:proxy_reformulate-xi}.
Consider the dataset with a single feature and nine datapoints given in table~\ref{tab:ex1}, and assume that the uncertainty set is 
\[
    \Xi = \left\{\boldsymbol \xi \in \mathbb{R}^{|\sets I| \times |\sets F|} \ : \ \Vert \boldsymbol \xi^i \Vert_\infty \leq 1 \quad \forall i \in \sets I \right\}.
\]
The only optimal solution of~\eqref{eq:single_setup} is given by the function~$\pi^\star(x)=\mathbb{I}(x\geq 6)$, which in the worst case misclassifies the two datapoints with feature values~$x=5$ and~$x=6$
However,~$\pi^\star(x)$ is not a feasible solution to~\eqref{eq:proxy_reformulate-xi}, as the fifth sample has~$\pi^\star(5) = 0$, but with perturbation~$\xi^5 = 1$,~$\pi^\star(5 + 1) = 1$. This violates the constraints in~\eqref{eq:proxy_reformulate-xi}, and thus~$\pi^\star(x)=\mathbb{I}(x\geq 6)$ is not a feasible solution to~\eqref{eq:proxy_reformulate-xi}. Thus, this example shows that all optimal solutions to~\eqref{eq:single_setup} may be infeasible in~\eqref{eq:proxy_reformulate-xi}.
\hfill \Halmos
\endproof

In the example from proposition~\eqref{prop:subset}, the optimal solution of~\eqref{eq:proxy_reformulate-xi}, and thus~\eqref{eq:proxy-xi}, is given by~$\bar \pi(x)=\mathbb{I}(x\geq 2)$, resulting in the missclassification of the three datapoints with~$x\in \{3,4,5\}$. We see that the two problems yield different optimal solutions, and that~$\bar \pi(x)$ is a more conservative solution that misclassifies more samples than~$\pi^\star(x)$.

Because problems~\eqref{eq:proxy-xi} and~\eqref{eq:proxy_reformulate-xi} are equivalent, by proposition~\ref{prop:subset}, problem~\eqref{eq:single_setup} considers no less than the set of robust solutions that~\eqref{eq:proxy-xi} considers. Thus, problem~\eqref{eq:single_setup} may lead to potentially less conservative solutions in comparison to~\eqref{eq:proxy-xi}, resulting in better performance as the example in the proof of proposition~\ref{prop:subset} shows:~\eqref{eq:single_setup} considers additional trees with more branching decisions and fewer misclassifications in both the nominal and worst cases. Indeed as uncertainty~$\epsilon$ grows, the only feasible tree for~\eqref{eq:proxy-xi} that can ensure that data samples always land at the same leaf no matter how the data is perturbed is a tree of depth zero (with no branching node).

\begin{table}
    \centering
    \begin{tabular}{c||c|c|c|c|c|c|c|c|c}
        x & 1 & 2 & 3 & 4 & 5 & 6 & 7 & 8 & 9 \\
        \hline
        y & 0 & 0 & 0 & 0 & 1 & 1 & 1 & 1 & 1
    \end{tabular}
    \caption{Example data.}
    \label{tab:ex1}
\end{table}

\section{Experiments}
\label{sec:experiments}
In this section, we evaluate our method on various datasets and across uncertainty set parameters. We assess the effect of robustness on accuracy under shifts, accuracy under no shifts (a.k.a. price of robustness)~\citep{bertsimas2004price}, sparsity, and computation times of our approach.

We refer to the Electronic Companion~\ref{appendix:experiment_bertsimas} for an empirical comparison of our method to the method of~\cite{Bertsimas2019RobustClassification} for learning robust trees, where we show that our method overall outperforms the method of~\cite{Bertsimas2019RobustClassification} in terms of worst-case and average-case accuracy and confirm empirically the theoretical observations made in section~\ref{sec:compare_robustness}.

\subsection{Setup and Instances}
We conduct all our experiments in Python 3.6 with Gurobi 9.0.1 as our MIO solver. All problems are solved on a single core of an Intel Xeon Processor 2640v4 and 4GB of memory with a time limit of 7200 seconds.
For instances that do not solve to optimality within the time limit, we return the best feasible tree found within the time limit and record the corresponding optimality gap reported by the solver. 

\subsubsection{Instances} Each instance of our experiments consists of a choice of uncertainty set parameters, a maximum depth of tree, and a dataset. For each instance's uncertainty set, we utilize the hypothesis testing framework as described in section~\eqref{sec:stat_connect}: for each~$f \in \mathcal{F}$, we choose the probability of certainty~$\rho^i_f$ by sampling from a normal distribution with a standard deviation of 0.2, keeping~$\rho^i_f$ the same across all~$i \in \mathcal{I}$ for each~$f$, and assuming no bounds on any non-categorical features. Note that if the number sampled from the normal distribution is greater than 1 (resp. less than 0), then the value of~$\rho^i_f$ is set to 1 (resp. 0). For the means of this normal distribution, we create instances with means of 0.6, 0.7, 0.8, and 0.9. An instance is also created for different budgets of uncertainty~$\epsilon$ by setting~$\lambda$ to be 0.5, 0.75, 0.85, 0.9, 0.95, 0.97, and 0.99. For every dataset and uncertainty set, we test with tree depths of 2, 3, 4, and 5. 

We evaluate on 12 publicly available datasets as listed in Table~\ref{tab:datasets}. Each dataset contains either integer-valued data, categorical data, or a mixture of both, as detailed in the table. We preprocess datasets with categorical data by one-hot encoding each categorical feature. 
The number of samples in the datasets range from~124 to~3196 and the number of features from~4 to~36. For each dataset, we randomly split it into 80\% training data and 20\% testing data once. In total, we experiment on 1536 instances (128 per dataset \revision{and 192 per value of~$\lambda$}) each with different uncertainty sets and maximum depths of tree.

\begin{table}[]
    \centering
    \begin{tabular}{|c|c|c|c|c|c|c|}
        \hline
         Name & Samples & Ints. & Cats. & $|\sets F|$ & Range of Ints. & Num. Cat. Values \\
         \hline
         balance-scale & 625 & 4 & 0 & 4 & 5 & N/A \\
         breast-cancer & 277 & 0 & 9 & 38 & N/A & 1-11 \\
         car-evaluation & 1728 & 5 & 1 & 6 & 4-55 & 1 \\
         hayes-roth & 132 & 4 & 0 & 4 & 3-4 & N/A \\
         house-votes-84 & 232 & 0 & 16 & 16 & N/A & 1 \\
         kr-vs-kp & 3196 & 0 & 36 & 38 & N/A & 1-3 \\
         monk1 & 124 & 6 & 0 & 6 & 2-4 & N/A \\
         monk2 & 169 & 6 & 0 & 6 & 2-4 & N/A \\
         monk3 & 122 & 6 & 0 & 6 & 2-4 & N/A \\
         soybean-small & 47 & 35 & 0 & 35 & 2-7 & N/A \\
         spect & 267 & 0 & 23 & 22 & N/A & 1 \\
         tic-tac-toe & 958 & 0 & 9 & 27 & N/A & 3 \\
         \hline
    \end{tabular}
    \caption{Datasets used from the UCI Machine Learning Repository \citep{Dua2017UCIRepository}. Columns from left to right: name of dataset, number of samples, number of integer features, number of categorical features, value of $|\sets F|$, range of integer features (if applicable), and number of categorical values (if applicable).}
    \label{tab:datasets}
\end{table}

\subsubsection{Generating Shifted Test Sets} \label{sec:generate_shifts}
To test our method's robustness against distribution shifts, we generate~5,000 different perturbed test sets for each instance. To create each perturbed test set, we independently perturb the original test data based on expected perturbations. That is, for the collection of~$q^i_{f}$ values for every~$f \in \mathcal{F}$ used to construct an uncertainty set based off of~\eqref{eq:hyptest}, we perturb each test set based on the symmetric geometric distribution described in~\eqref{eq:symmetric_geometric}.

In order to measure the robustness of our method against unexpected shifts of the data, we repeat the same process of generating shifted test sets for each instance but with \emph{unexpected perturbations}: perturbing the test data using values of~$\rho^i_f$ different than what we gave our model. We shift each~$\rho^i_f$ value down 0.2, then perturb our test data in 5,000 different ways based on these new values of~$\rho^i_f$. We do the same procedure but with~$\rho^i_f$ shifted down by 0.1 and up by 0.1. In a similar fashion, we also uniformly sample a new~$\rho^i_f$ value for each feature in a neighborhood of radius 0.05 of the original expected~$\rho^i_f$ value, and perturb the test data in 5,000 different ways with the new~$\rho^i_f$ values. We do the same procedure for the radii of 0.1, 0.15, and 0.2 for the neighborhoods.

\revision{Note that we also add a small regularization parameter to our robust model to discourage redundant branching on features and thresholds. Specifically, the objective function in problem~\eqref{eq:additional_cuts_formulation} is slightly modified to:
\[
    \sum_{i\in\mathcal{I}}t_i - \frac{1}{2^d}\sum_{n\in\mathcal{N}}\sum_{f\in\mathcal{F}}\sum_{\theta \in \Theta(f)}b_{nf\theta}.
\]
As the regularization term is always smaller than 1, the accuracy under worst-case distribution shift is the same between the above problem and problem~\eqref{eq:additional_cuts_formulation}, so a tree optimal in the above problem is also optimal in~\eqref{eq:additional_cuts_formulation}. We do not add any more regularization to our robust model otherwise, as we will empirically show that adding robustness itself has a regularizing (sparsity-promoting) effect in section~\ref{sec:sparsity}. }

\subsubsection{Learning Non-Robust Trees} 
\label{sec:learning_other_trees}
For comparison, we use our method to create a non-robust optimal tree by setting the budget of uncertainty to 0 (i.e.,~$\lambda=1$), and tune a regularization parameter for the non-robust tree using a validation set from the training set. The regularization term penalizes the number of branching nodes to yield the modified objective in problem~\eqref{eq:additional_cuts_formulation}:
\[
    \max_{\mathbf{b}, \mathbf{w},\mathbf{t}}\; R\sum_{i\in\mathcal{I}}t_i - (1-R)\sum_{i\in\mathcal{I}}\sum_{n\in\mathcal{N}}\sum_{\Theta(f)}b_{nf\theta},
\]
where~$R \in [0,1]$ is the tuned regularization parameter. To tune the value~$R$, we first randomly hold out 20\% of the training set into a validation set. Then, we select various values of~$R$ from the set~$\{0.6,0.7,0.8,0.9,0.95,1\}$ and learn a tree using our method with the uncertainty set~$\{\boldsymbol 0\}$ and the same specifications as learning our robust tree but instead with a time limit of~1~hour. We find the accuracy of the learned tree on the held-out validation set, and select~$R$ with the best accuracy. We then create the non-robust optimal tree using the entire training set and tuned~$R$ value.

\subsection{Effect of Robustness}
We now examine our method's robustness to distribution shifts across different uncertainty sets in comparison to a non-robust model.

\subsubsection{Accuracy Under Distribution Shifts}
\label{sec:acc_tradeoffs}
\begin{figure}[t!]
\centering
\begin{subfigure}[b]{0.49\textwidth}
\centering
\includegraphics[width=\textwidth]{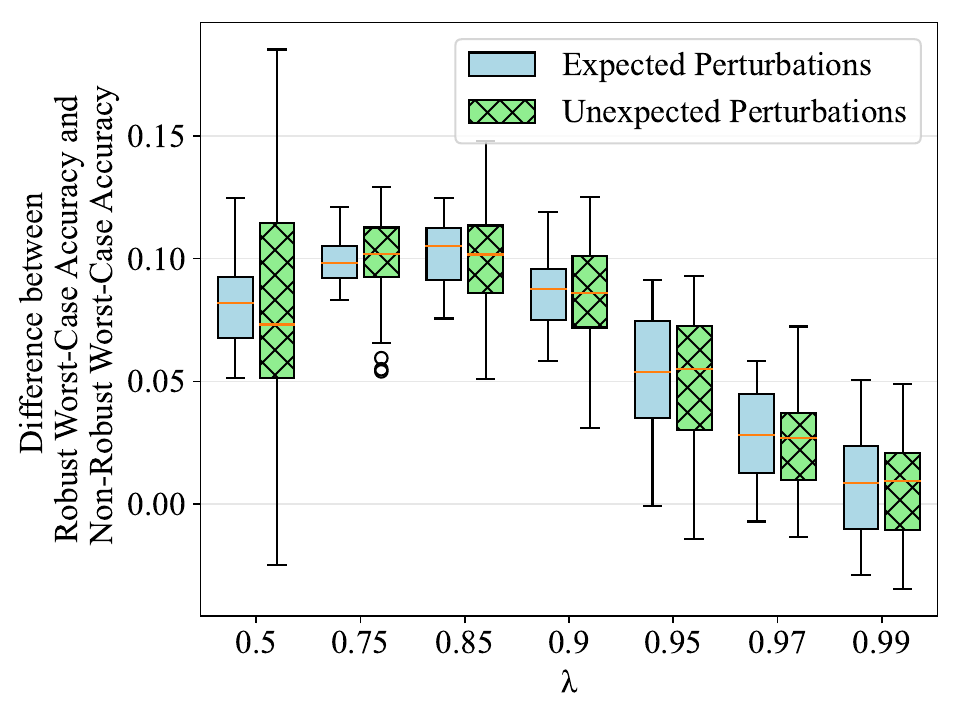}
\caption{}
\end{subfigure}
\begin{subfigure}[b]{0.49\textwidth}
\centering
\includegraphics[width=\textwidth]{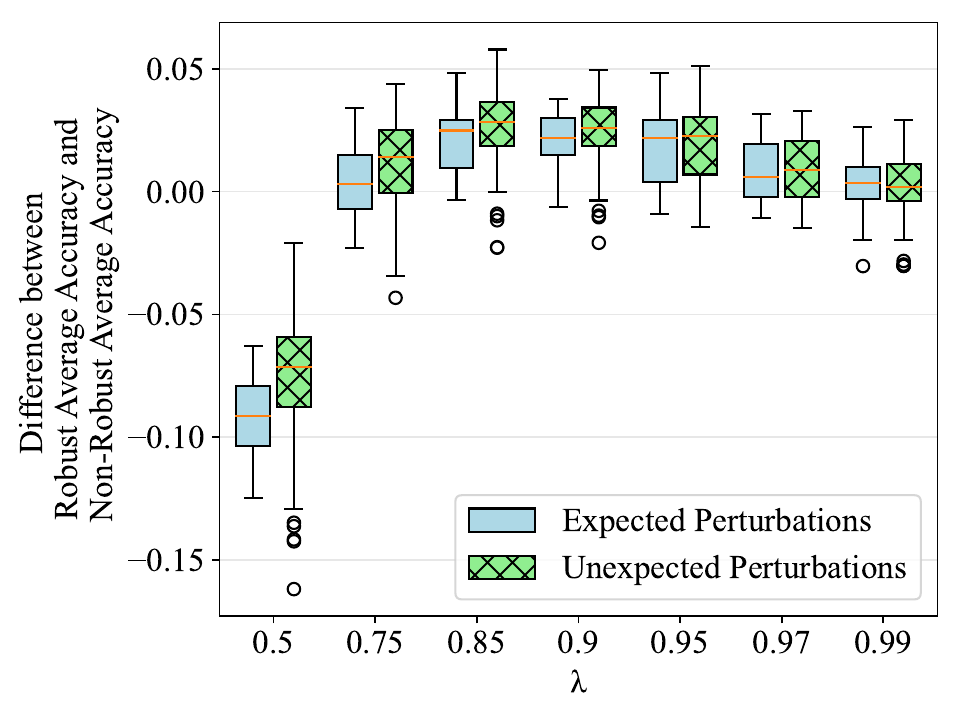}
\caption{}
\end{subfigure}
\caption{Companion figure to section~\ref{sec:acc_tradeoffs}. Distributions across problem instances of the gain in (a) worst-case accuracy and (b) average accuracy from using a robust tree versus a non-robust, regularized tree in dependence of~$\lambda$ under both expected and unexpected perturbations in the test data. The parameter~$\lambda$ varies between 0.5 (more robust) and 0.99 (less robust).}\label{fig:boxworst_avg}
\end{figure}

For each instance, we measure the worst-case (lowest) and average accuracy across all perturbed test sets. Our results are summarized in Figure~\ref{fig:boxworst_avg}, which shows across all instances the distributions of the difference in worst-case and average accuracies between the learned robust and non-robust trees. From the figure, we see that the robust model in general has higher worst-case and average-case accuracies than the non-robust model when there are distribution shifts in the data. We also see that there is a range of values of~$\lambda$ (namely between 0.75 and 0.9) that perform well over other values of~$\lambda$ in terms of both worst-case and average case accuracy. This shows us that if the budget of uncertainty is too small, then we do not allow enough room to hedge against distribution shifts in our uncertainty set. But if the budget of uncertainty is too large, then we become over-conservative and lose some accuracy in both expected and unexpected perturbations of our test data. We also see that there is little difference between the gains in accuracy in instances where the perturbation of our data is as we expect versus when the perturbation is not as we expect. This indicates that even if we misspecify our model, we still obtain a classification tree robust to distribution shifts within a reasonable range of the expected shift. Overall, we see that an important factor in determining the performance of our model is the budget of uncertainty, which can be conveniently tuned to create an effective robust tree. 

\subsubsection{Price of Robustness}
\label{sec:price_robustness}
\begin{figure}[t!]
\centering
\begin{subfigure}[b]{0.49\textwidth}
\centering
\includegraphics[width=\textwidth]{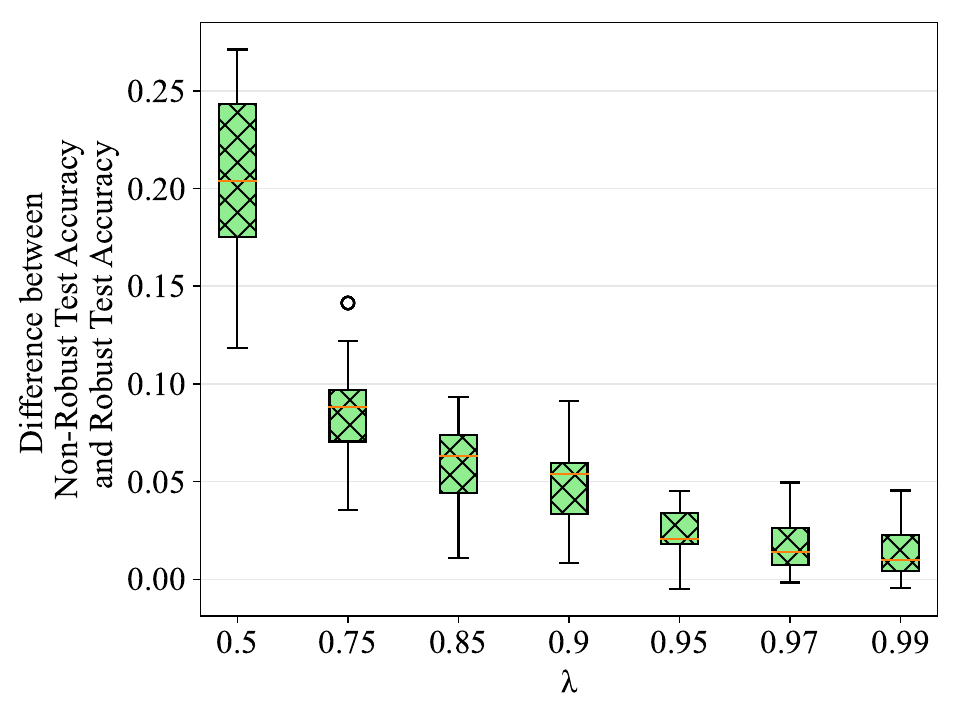}
\vspace{-\baselineskip}
\caption{}
\label{fig:boxprice}
\end{subfigure}
\hfill
\begin{subfigure}[b]{0.49\textwidth}
\centering
\includegraphics[width=\textwidth]{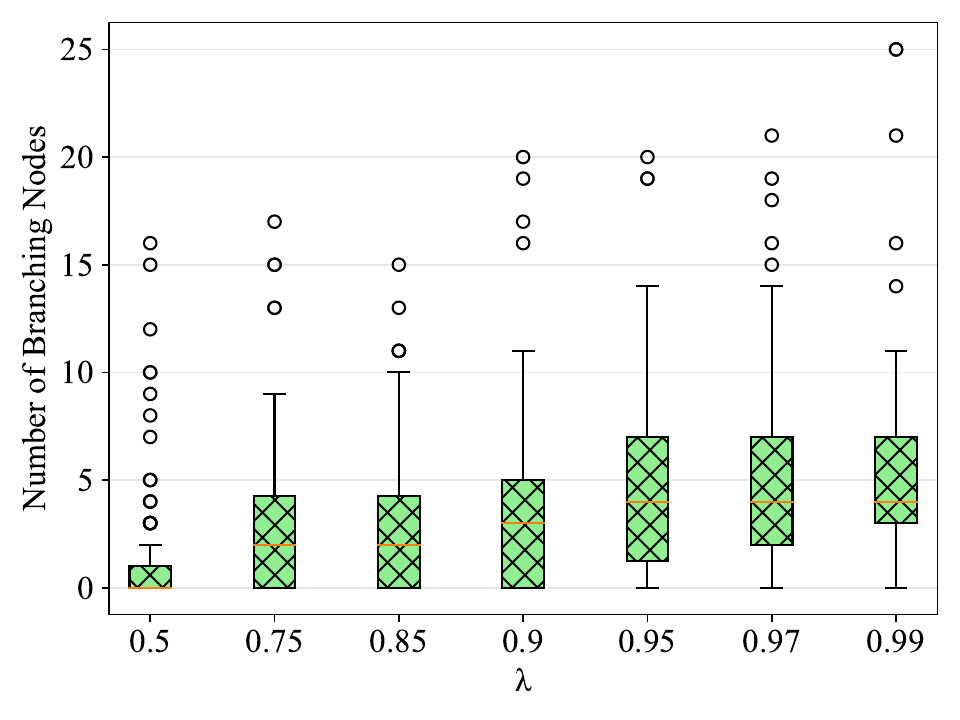}
\vspace{-\baselineskip}
\caption{}
\label{fig:num_branching}
\end{subfigure}
\caption{(a) Companion figure to section~\ref{sec:price_robustness}. Distribution across problem instances of the difference in nominal test accuracy between a non-robust, regularized tree and a robust tree in dependence of~$\lambda$. The parameter~$\lambda$ varies between 0.5 (more robust) and 0.99 (less robust). (b) Companion figure to section~\ref{sec:sparsity}. Distributions of the number of branching nodes across all instances with the same~$\lambda$. The parameter~$\lambda$ varies between 0.5 (more robust) and 0.99 (less robust).}
\label{fig:boxprice_num_branching}
\end{figure}

We also measure the decrease in accuracy from using a robust tree versus a non-robust tree under the case of no distribution shift in the test set, i.e., the price of robustness \citep{bertsimas2004price}, and summarize this metric in Figure~\ref{fig:boxprice}. From the figure, we observe that in the range of~$\lambda$ values that perform well in terms of accuracy (i.e., between~$0.75$ and~$0.9$, see section~\ref{sec:acc_tradeoffs}), we have an average price of robustness between~$0.05$ and~$0.1$. As the level of robustness increases (with a decreasing~$\lambda$), the higher the price of robustness. This is expected with our model because a larger budget of uncertainty leads to larger deviations away from the nominal test set considered, and thus the worst-case distribution shift may look very different from the nominal test set if more drastic shifts are considered. 

\subsubsection{The Effect of Robustness on Sparsity}
\label{sec:sparsity}
To evaluate the sparsity of our model in relation to the robustness of the tree, we measure the number of branching nodes in each instance. Our results are summarized in Figure~\ref{fig:num_branching}. Overall, as the size of the uncertainty set increases with a smaller~$\lambda$, the number of branching nodes decreases. Namely, the median number of branching nodes is 4 for instances with~$\lambda \in \{0.95, 0.97, 0.99\}$, 3 for instances with~$\lambda = 0.9$, 2 for instances with~$\lambda \in \{0.75, 0.85\}$, and 0 for instances with~$\lambda =0.5$\revision{.} This regularizing behavior of robustness is expected with our model: with more branching nodes in the learned model, the more opportunities there are for a low-cost perturbation of each sample in a given tree, yielding a lower worst-case accuracy. Thus, as the number of perturbations allowed expand, the number of branching nodes in the learned model decreases to yield a more favorable worst-case accuracy.

\subsubsection{Computational Times}
\label{sec:runtimes}
\begin{figure}[t]
\centering
\includegraphics[width=0.49\textwidth]{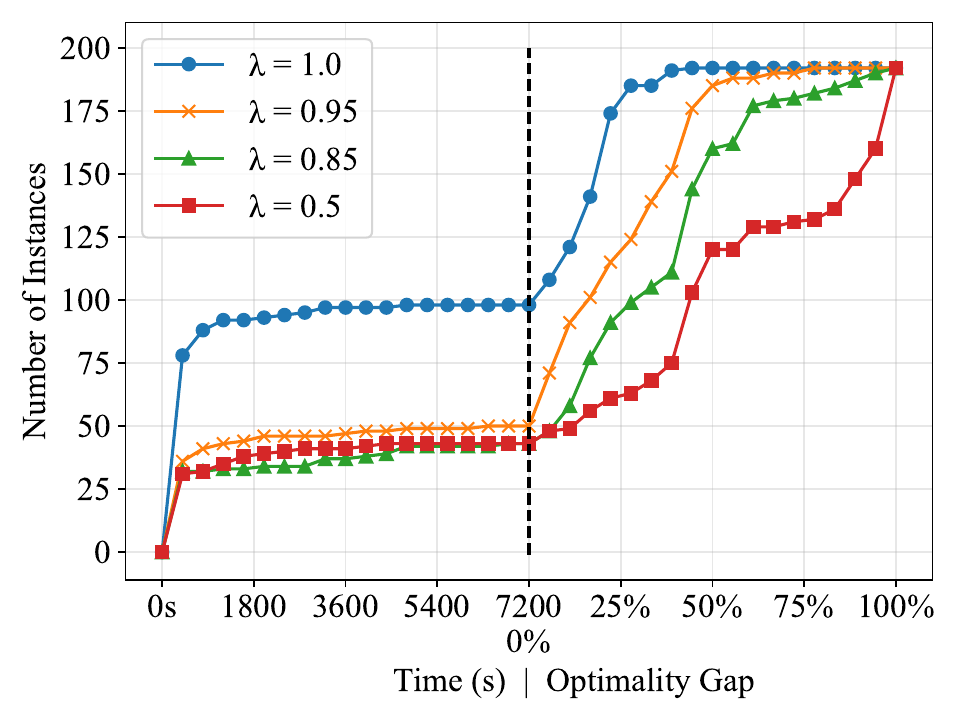}
\caption{Companion figure to section~\ref{sec:runtimes}. \revision{
Cumulative number of problems solved 
to optimality for several values of~$\lambda$ on the `Time' axis, and cumulative number of problems with MIO optimality gap no larger than the given value on the `Optimality Gap' axis.} The case of~$\lambda = 1.0$ corresponds to the non-robust, regularized tree.}
\label{fig:runtime}
\end{figure}

We summarize the computation times across all instances in Figure~\ref{fig:runtime}.
For any fixed~$\lambda$, the variations in computation times and optimality gaps across instances are due to differences in the maximum depth of tree, the number of data samples, the number of features, and the range of values for each feature within each instance. In general, a larger uncertainty set (smaller $\lambda$) leads to a larger optimality gap for a fixed time limit, as there are more constraints to add to the master problem to reach optimality. In particular, the number of instances that have an optimality gap larger than 50\% are 72, 32, 7, and 0 for $\lambda = 0.5$,~$ 0.85$,~$0.95$, and~$1$, respectively.
We also observe that there is a large gap in number of instances solved fully to optimality within the time limit between the non-robust instances and the robust instances, with 48 more instances solved to optimality in the non-robust case in comparison to the robust case with~$\lambda = 0.95$. In addition, the number of robust instances able to be solved to optimality within the time limit are fairly close, with only 7 instances between the~$\lambda = 0.95$ case and the~$\lambda = 0.85$ and~$\lambda = 0.5$ cases.

\section{Acknowledgments}

N.\ Justin is funded in part by the National Science Foundation Graduate Research Fellowship Program. P.\ Vayanos and N.\ Justin are funded in part by the National Science Foundation under CAREER grant 2046230. A.\ Gómez is funded in part by the National Science Foundation under grants 1930582 and 2152777. P.\ Vayanos and A.\ Gómez are funded in part by the National Science Foundation under NRT grant 2346058. They are grateful for the support.

\bibliographystyle{informs2014} 
\bibliography{references_final.bib}

\newpage

\ECSwitch

\ECHead{E-Companion}

\revision{
\section{Connection to Wasserstein Distributionally Robust Optimization}
\label{appendix:dro}

In this section, we present a Wasserstein distributionally robust formulation for learning optimal classification trees robust to distribution shifts, and discuss the relationship between this model and the robust optimization model that we study in this paper. For general details on the mathematical relationship between distributionally robust optimization and robust optimization, we refer to corollary 2(iii) in~\cite{gao2023distributionally}.

We first define the Wasserstein distance for data with discrete features. Let $X$ be a random vector from the space of covariates~$\mathcal{X} \subseteq \mathbb{Z}^{|\mathcal{F}|}$, and~$Y$ be a random variable denoting the label from the set~$\mathcal{K}$. Let the set of all probability measures for $(X,Y)$ be in set $\mathcal{P}$, and we will say that the training data is drawn from a distribution $\mathbb{P}^0 \in \mathcal{P}$. The empirical distribution of our covariates is defined as 
\[
    \hat{\mathbb{P}}_X(\mathbf{x}) := \frac{1}{|\mathcal{I}|}\sum_{i \in \mathcal{I}}\mathbb{I}\left[\mathbf{x} = \mathbf{x}^i\right].
\]
Let~$\mathbb{Q} \in \mathcal{P}$ be the realization of the distribution of data after distribution shift, and~$\mathbb{Q}_X$ denote the marginal distribution of~$\mathbb{Q}$ on~$X$.

To measure the distance between~$\mathbb{Q}_X$ and the empirical distribution~$\hat{\mathbb{P}}_X$, we use the Wasserstein distance, which over a discrete space is defined as
\begin{align*}
    W(\hat{\mathbb{P}}_X, \mathbb{Q}_X) := \inf_{\mathbb{M} \in \mathcal{P}_X^2} & \sum_{\mathbf{x} \in \mathcal{X}}\sum_{\mathbf{x}' \in \mathcal{X}} d(\mathbf{x}, \mathbf{x}') \mathbb{M}(\mathbf{x}, \mathbf{x}') \\
     \text{s.t. }&\sum_{\mathbf{x}' \in \mathcal{X}} \mathbb{M}(\mathbf{x}', \mathbf{x}) = \mathbb{Q}_X(\mathbf{x}) \\
     & \sum_{\mathbf{x}' \in \mathcal{X}} \mathbb{M}(\mathbf{x}, \mathbf{x}') = \hat{\mathbb{P}}_X(\mathbf{x})
\end{align*}
for a distance metric $d$ on $\mathcal{X}$ and $\mathcal{P}_X^2$ the set of all joint probability measures for $(X,X)$. To create an analogous ambiguity set as the uncertainty set defined in~\eqref{eq:generalUncertainty}, we define the following distance metric between points indexed in the set $\mathcal{I}$ and any $\bar{\mathbf{x}} \in \mathcal{X}$:
\[
    d(\mathbf{x}^i, \bar{\mathbf{x}}) := \sum_{f \in \mathcal{F}} \gamma^i_f\left\vert x^i_f - \bar{x}_f \right\vert.
\]
Then, letting $\mathbb{Q}^i_X(\mathbf{x})$ be the distribution of $\mathbb{M}$ conditioned on $\mathbf{x}^i$, we can represent our Wasserstein distance as
\begin{equation}\label{eq:wasserstein}
    W(\hat{\mathbb{P}}_X, \mathbb{Q}_X) = \sum_{i \in \mathcal{I}}\sum_{\bar{\mathbf{x}} \in \mathcal{X}}\sum_{f \in \mathcal{F}} \gamma^i_f \left\vert x^i_f - \bar{x}_f\right\vert\mathbb{Q}^i_X(\bar{\mathbf{x}}).
\end{equation}

Recall that we are interested in the setting where there is only a shift in the covariates. Thus, we impose the following assumption on the distribution shifts we consider:
\begin{assumption}\label{as:no-label-shift}
    Let the observed data on individual~$i\in\mathcal{I}$ be $(\hat{X}^i, \hat{Y}^i) \sim \mathbb{P}^0$, where individual~$i$ in the unknown shifted distribution has data $(X^i,Y^i) \sim \mathbb{Q}$ for unknown $\mathbb{Q} \in \mathcal{P}(\Omega)$. We assume that the shifted distribution $\mathbb{Q}$ has the property $Y^i = \hat{Y}^i$ for all $i \in \mathcal{I}$.
\end{assumption}
Using this assumption and the definition of the Wasserstein distance defined in~\eqref{eq:wasserstein}, we then restrict the set of all possible distributions via a Wasserstein ball with radius~$\epsilon \geq 0$:
\begin{equation}\label{eq:w_ball}
    \mathcal{B}(\epsilon) := \{\mathbb{Q} \in \mathcal{P} \ : \ W(\mathbb{Q}_X, \hat{\mathbb{P}}_X) \leq \epsilon, \ \mathbb{Q} \text{ satisfies Assumption~\ref{as:no-label-shift}}\}.
\end{equation}
Then, the analogous distributionally robust problem to~\eqref{eq:single_setup} is
\begin{equation} \label{eq:dro_problem}
    \max_{\pi \in \Pi_d}\; \inf_{\mathbb{Q} \in \mathcal{B}(\epsilon)} \;\; |\mathcal{I}|\mathbb{E}_{\mathbb{Q}}\left[\mathbb{I}\left[ \pi(X) = Y \right]\right],
\end{equation}
where we multiply the objective by $|\mathcal{I}|$ to make the objective the expected number of training samples that are correctly classified, making the objective similar to~\eqref{eq:single_setup}.

Consider adding a constraint in~$\mathcal{B}(\epsilon)$ that $\mathbb{Q}^i_X(\mathbf{x}) \in \{0, 1\}$ for all $\mathbf{x} \in \mathcal{X}$. For any $\mathbb{Q} \in \mathcal{B}(\epsilon)$ that satisfies this constraint, there exists exactly one $\bar{\mathbf{x}}^i \in \mathcal{X}$ such that $\mathbb{Q}^i_X(\bar{\mathbf{x}}^i) = 1$ for each $i \in \mathcal{I}$. Define a set~$\bar{\mathcal{X}}(\epsilon) \subseteq \mathbb{Z}^{|\mathcal{F}|}$ that collects all of these~$\bar{\mathbf{x}}^i$ values:
\[
    \bar{\mathcal{X}}(\epsilon) := \left\{\bar{\mathbf{x}} \in \mathbb{Z}^{|\mathcal{I}| \times |\mathcal{F}|} \ : \ \exists \mathbb{Q} \in \mathcal{B}(\epsilon) \ \text{s.t.} \ \mathbb{Q}^i_X(\bar{\mathbf{x}}^i) = 1 \ \forall i \in \mathcal{I} \right\}.
\]
By estimating the expectation in~\eqref{eq:dro_problem} via the training sample average, we can represent~\eqref{eq:dro_problem} under the additional constraint that~$\mathbb{Q}^i_X(\mathbf{x}) \in \{0, 1\}$ as the following problem:
\begin{equation}\label{eq:dro_to_ro}
    \max_{\pi \in \Pi_d}\; \min_{\bar{\mathbf{x}} \in \bar{\mathcal{X}}(\epsilon)} \;\; \sum_{i \in \mathcal{I}}\mathbb{I}\left[ \pi(\bar{\mathbf{x}}^i) = y^i \right],
\end{equation}
which with a change of variables to~$\boldsymbol{\xi}^i := \bar{\mathbf{x}}^i - \mathbf{x}^i$, becomes equivalent to~\eqref{eq:single_setup}.

One interpretation of the Wasserstein distributionally robust problem~\eqref{eq:dro_problem} is that it allows for probabilistic realizations of our data, as represented by the value of~$\mathbb{Q}$. In the approach presented in our paper~\eqref{eq:single_setup}, we hedge against possible realizations of our data under the distribution seen in testing, and do not allow for these realizations to be probabilistic. A key observation is that, as shown by the problem~\eqref{eq:dro_to_ro}, the set of worst-case realizations explored in the robust approach in~\eqref{eq:single_setup} is a subset of the worst-case realizations in the distributionally robust approach~\eqref{eq:dro_problem}. This means that~\eqref{eq:single_setup} may generate less conservative solutions to solving~\eqref{eq:dro_problem}, as~\eqref{eq:single_setup} only hedges against possible realizations of the data. Lastly, we note that since~\eqref{eq:single_setup} contains additional constraints in the inner minimization problem in comparison to~\eqref{eq:dro_problem}, the objective value of~\eqref{eq:dro_problem} must be no more than the objective value of~\eqref{eq:single_setup}. We formalize the relationship between the robust optimization problem we examine in this paper~\eqref{eq:single_setup} and the distributionally robust optimization version of this problem~\eqref{eq:dro_problem} in the following proposition.

\begin{proposition}
    The set of feasible solutions to the robust optimization problem~\eqref{eq:single_setup} are a subset of feasible solutions to the distributionally robust optimization problem~\eqref{eq:dro_problem}, and the objective value of~\eqref{eq:single_setup} is no less than the objective value of~\eqref{eq:dro_problem}.
\end{proposition}

}

\section{Iterating over Paths in Problem~\eqref{eq:subproblem}} \label{appendix_sec:paths}
We will now show why the assumption that the cut-set~$\mathbf{q}^i$ incident to the path of the data sample after perturbation minimizes~\eqref{eq:subproblem} for given~$\boldsymbol{\xi}$ through the following proposition.
\begin{proposition}\label{prop:optimal}
For optimal solutions of~\eqref{eq:subproblem}, one of the following two statements holds for each sample~$i \in \mathcal{I}$
\begin{enumerate}
\item The source set of a minimum cut is~$\{s\}$, that is,~$q^i_{s, 1}=1$ and~$q^i_{n, l(n)}=q^i_{n, r(n)}=0$ for all~$n\in \sets N$. In this case~$t^i=1$.
\item The source set of a minimum cut is a path from~$s$ to a prediction node with a label other than~$y^i$, that is,~$\bar w_{n,y^i}=0$. In this case~$t^i=0$.
\end{enumerate}
\end{proposition}

\proof{Proof of Proposition~\ref{prop:optimal}}
First observe that setting~$q^i_{s, 1}=1$ and all remaining variables to zero indeed satisfies constraints~\eqref{eq:mincutset_left}-\eqref{eq:mincutset_sink}, and the associated objective value is~$t^i=1$. Suppose the first statement in the proposition does not hold and this solution is not optimal. Since, for all values of~$\boldsymbol{\xi}^i$, the arc capacities are either~$0$ or~$1$, it follows that the objective values corresponding to optimal solutions is~$t^i=0$. 
Note that if a prediction node~$n$ with label~$y^i$ ($\bar w_{n,y^i}=1$) is in the cut ($q_{n,t}^i=1)$, then~$t^i\geq 1$, and no such solutions can be optimal. 

If a branching node~$n$  ($\bar b_{nf\theta}=1$ for some feature~$f$ and level~$\theta$) is in the cut but none of its descendants are ($q_{n,l(n)}^i=q_{n,r(n)}^i=1$), then~$\mathbb{I}\left[x^i_f + \xi^i_f \leq \theta\right]q^i_{n, l(n)}+\mathbb{I}\left[x^i_f + \xi^i_f \geq 1+\theta\right]q^i_{n, r(n)}=1$ for all values of~$\xi_f^i$. Thus in this case~$t^i\geq 1$ and no such solutions can be optimal. If a branching node is in the cut and both of its descendants are as well ($q_{n,l(n)}^i=q_{n,r(n)}^i=0$), then depending on the value of~$\boldsymbol{\xi}^i$, either~$\mathbb{I}\left[x^i_f + \xi^i_f \leq \theta\right]=0$ or~$\mathbb{I}\left[x^i_f + \xi^i_f \geq 1+\theta\right]=0$. In the first case, one may set~$q^i_{n, l(n)}=1$ (and set additional cutset variables on the left subtree to zero), recovering a solution with the same (or less) cost; the second case is analogous. Therefore, we find that if~$t^i=0$, then there exists an optimal solution that is a path. \hfill \Halmos
\endproof

Due to proposition~\ref{prop:optimal}, we only need to solve \eqref{eq:cost_perturb} over all possible paths~$\mathcal{O}(|\sets L|)$ for each sample to solve the subproblem and obtain a violated constraint~\eqref{eq:violated_constraint} to add back to the main problem.

\section{Computation Times Improvement from Adding Constraints~\eqref{eq:violated_constraint_single}}
\label{sec:runtimes_nocut_comparison}
\begin{figure}[t]
\centering
\includegraphics[width=0.49\textwidth]{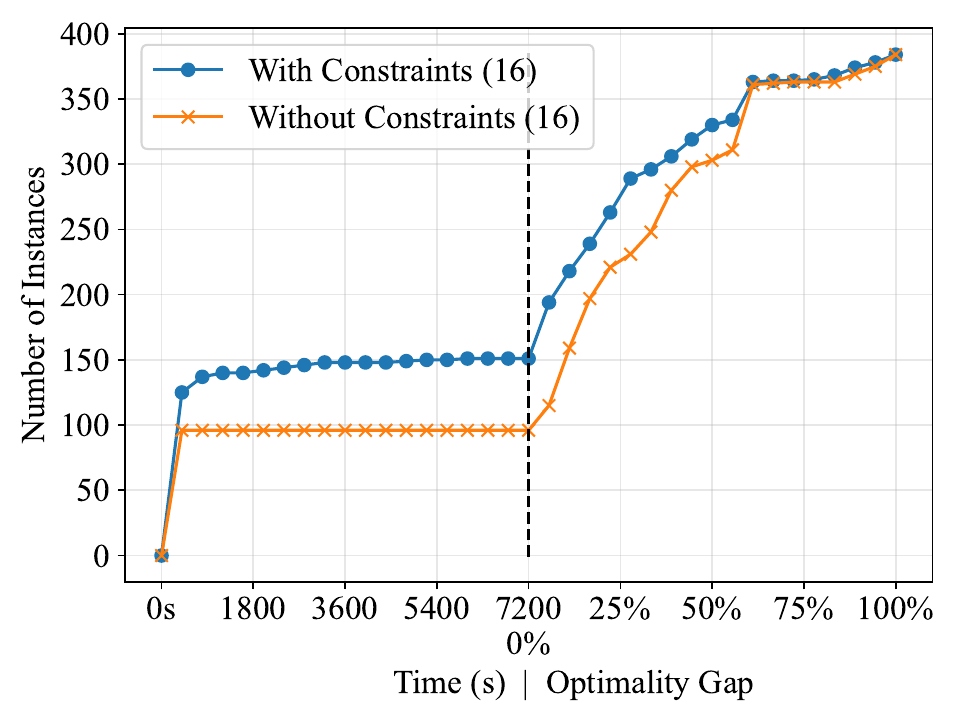}
\caption{Companion figure to section~\ref{sec:runtimes_nocut_comparison}\revision{. N}umber of \texttt{monk1}, \texttt{monk2}, and \texttt{monk3} instances solved \revision{to optimality on the `Time' axis and the number of instances with MIO optimality gap no larger than each given value at the time limit on the `Optimality Gap' axis}. The blue line corresponds to the algorithm presented in this paper where constraints~\eqref{eq:violated_constraint_single} are added at each subproblem, and the orange line is the same algorithm where constraints~\eqref{eq:violated_constraint_single} are not added.}
\label{fig:runtime_nocut}
\end{figure}

We add additional constraints of the form~\eqref{eq:violated_constraint_single} to our model in order to improve on computation times. We test the magnitude of this improvement by comparing our algorithm that adds constraints~\eqref{eq:violated_constraint_single} against the algorithm that does not add constraints~\eqref{eq:violated_constraint_single} at each subproblem. We run such algorithms on the instances with the \texttt{monk1}, \texttt{monk2}, and \texttt{monk3} datasets, and summarize the computation times and optimality gaps in Figure~\ref{fig:runtime_nocut}. We observe that 96 out of the 384 instances are solved within the time limit on the algorithm without added constraints~\eqref{eq:violated_constraint_single}. Our algorithm with additional constraints~\eqref{eq:violated_constraint_single} is able to solve the same number of instances within 32 seconds, corresponding to a speedup of~$\frac{7200}{32} = 225$ times.

\section{Variants and Extensions}
\label{sec:extensions}
In this section, we describe how our model and solution approach can be adapted to handle bounded integer data, one-sided distribution shifts, distribution shifts on categorical features, and distribution shifts on mixed datasets.

\subsection{Calibration of Uncertainty Set Parameters for Bounded Integer Data}
\label{sec:bounded}
In many applications, there exist known bounds on the values of the integer features. There may be a bound on only one end of an integer feature (e.g., age, income) or on both ends (e.g., integer rating on a bounded scale, binary features).
Here, we discuss how to tune the uncertainty set parameters for datasets involving bounded integer features with known bounds.

\subsubsection{Integer Features with One-Sided Bounds}
\label{sec:one_sided_bound}

We first consider the case where all covariates admit only a one-sided bound. Without loss of generality, we assume that all bounds are lower bounds, and we denote the lower bound on feature~$f$ by~$L_f$ (a symmetric argument can be made if all features are upper bounded instead).

To tune the hyperparameters of~\eqref{eq:generalUncertainty}, we assume that the probability of certainty in the nominal value of feature~$f$ for sample~$i$, $\rho^i_f$, satisfies $0 < \rho^i_f < 1$. Under this assumption, the perturbation~$\xi^i_f$ is distributed as the truncated symmetric geometric distribution conditioned on~$x^i_f + \xi^i_f \geq L_f$. That is, for $\zeta^i_f$ the realization of the perturbation of~$x^i_f$, the probability that~$x^i_f$ is perturbed by~$\xi^i_f$ is
\begin{equation} \label{eq:bound_prob}
    \mathbb{P}(\xi^i_f = \zeta^i_f) := \rho^i_f (r^i_f)^{|\zeta^i_f|}
\end{equation}
for~$0 < r^i_f \leq 1$ some constant that makes~$\mathbb{P}$ a distribution over the support of all $\xi^i_f$ that satisfy~$x^i_f + \xi^i_f \geq L_f$.
To find the value of~$r^i_f$ for~\eqref{eq:bound_prob}, we utilize the following lemma. 
\begin{lemma} \label{lem:one_side_bound}
    For $0 < \rho^i_f < 1$, there exists a real-valued solution $r^i_f$ to 
    \begin{equation}\label{eq:lb_finitesum}
         \rho^i_f(r^i_f)^{\revision{x^i_f}-L+1} - (\rho^i_f + 1)r^i_f + 1 - \rho^i_f = 0
    \end{equation}
    such that $0 < r^i_f < 1$.
    Furthermore, when used in~\eqref{eq:bound_prob}, $r^i_f$ defines a valid distribution over the support of all $\xi^i_f\in \mathbb{Z}$ that satisfy~$x^i_f + \xi^i_f \geq L_f$.
\end{lemma}
\begin{proof}{Proof of Lemma~\ref{lem:one_side_bound}}
    To find an~$r^i_f$ that makes~\eqref{eq:bound_prob} a probability mass function, we solve the infinite polynomial equation
    \begin{equation} \label{eq:lb_infintesum}
        \sum_{\xi^i_f = L_f-x^i_f}^\infty \rho^i_f (r^i_f)^{|\xi^i_f|} = 1,
    \end{equation}
    and choose the solutions such that $0 < r^i_f < 1$.
    We can regroup terms in the left hand side of the above sum to get
    \begin{equation*}
        \rho^i_f + 2\sum_{\xi^i_f = 1}^{x^i_f - L_f} \rho^i_f (r^i_f)^{\xi^i_f} +  \sum_{\xi^i_f = x^i_f - L_f + 1}^\infty \rho^i_f (r^i_f)^{\xi^i_f} = 1.
    \end{equation*}
    Then, using the closed form of both a finite and infinite geometric series, an equivalent finite polynomial equation to solve problem~\eqref{eq:lb_infintesum} is
    \begin{align*}
        \rho^i_f + 2\left(\frac{\rho^i_fr^i_f\left(1 - (r^i_f)^{x^i_f - L_f}\right)}{1 - r^i_f}\right) +  \left(\frac{\rho^i_f(r^i_f)^{x^i_f - L_f}}{1 - r^i_f}\right) &= 1 \\
        \Leftrightarrow\;\rho^i_f - \rho^i_fr^i_f + 2\rho^i_fr^i_f - 2\rho^i_f(r^i_f)^{x^i_f - L_f+1} + \rho^i_f(r^i_f)^{x^i_f - L_f + 1} &= 1 - r^i_f \\
        \Leftrightarrow\; \rho^i_f(r^i_f)^{\revision{x^i_f}-L+1} - (\rho^i_f + 1)r^i_f + 1 - \rho^i_f &= 0
    \end{align*}
    which is exactly~\eqref{eq:lb_finitesum}.

    Now we show that a solution $0 < r^i_f < 1$ exists in the above equation. Let
    \begin{equation*}
        f(r^i_f) := \rho^i_f(r^i_f)^{\revision{x^i_f}-L+1} - (\rho^i_f + 1)r^i_f + 1 - \rho^i_f,
    \end{equation*} i.e. the function on the left-hand side of~\eqref{eq:lb_finitesum} that we wish to find real roots of. Thus,~$f(0) = 1 - \rho^i_f$ and~$f(1) = -\rho^i_f$. Since~$0 < \rho^i_f < 1$,~$f(0) > 0$, and $f(1) < 0$, by Bolzano's Theorem, there exists a solution~$r^i_f$ to~\eqref{eq:lb_finitesum} such that~$0 < r^i_f < 1$. \hfill \Halmos
\end{proof}
Standard root-finding approaches (e.g., the bisection method) can be used to solve~\eqref{eq:lb_finitesum} in order to find~$r^i_f$.
Once $r^i_f$ is found, we set up a hypothesis test with threshold~$\lambda$ of the form in a similar fashion as done in section~\ref{sec:stat_connect}, which yields the following condition on~$\xi_f^i$:
\begin{equation} \label{eq:bounded_test}
    \frac{\prod_{i \in \mathcal{I}} \prod_{f \in \mathcal{F}}\rho^i_f(r^i_f)^{|\xi^i_f|}}{\prod_{i \in \mathcal{I}} \prod_{f \in \mathcal{F}} \rho^i_f} \geq \lambda^{|\mathcal{I}|},
\end{equation}
reducing down to
\begin{equation} \label{eq:bounded_test_sum}
    \sum_{i \in \mathcal{I}}\sum_{f \in \mathcal{F}} |\xi^i_f|\log\left(\frac{1}{r^i_f}\right)\leq - |\mathcal{I}| \log \lambda.
\end{equation}
Therefore, the tuned hyperparameters for this case are~$\gamma^i_f = \log \frac{1}{r^i_f}$ and~$\epsilon = - |\mathcal{I}| \log \lambda$.

\subsubsection{Integer Features with Both Upper and Lower Bounds}
\label{sec:two_sided_bounds}
We now assume that there is both a lower bound~$L_f$ and an upper bound~$U_f$ on all features~$f$.
As is the case with the one-sided bounded features, we set up a hypothesis testing framework to tune the hyperparameters. We assume that $\frac{1}{U_f - L_f + 1} \leq \rho^i_f \leq 1$ for all $i \in \sets{I}$ and $f \in \sets{F}$, and utilize the same truncated symmetric geometric distribution characterized by~\eqref{eq:bound_prob} with support~$[L_f - x^i_f, U_f - x^i_f]$. Note that we place the lower bound $\frac{1}{U_f - L_f + 1}$ on $\rho^i_f$ as assuming $\rho^i_f = \frac{1}{U_f - L_f + 1}$ under maximal entropy makes the perturbation of feature $f$ at sample $i$ uniformly distributed, meaning that there is maximal uncertainty on the value of $x^i_f + \xi^i_f$. For $\rho^i_f > \frac{1}{U_f - L_f + 1}$, we find an $ 0 < r^i_f < 1$ for each sample $i$ by the following lemma:
\begin{lemma} \label{lem:two_side_bound}
     For $\frac{1}{U_f - L_f + 1} < \rho^i_f < 1$, there exists a real-valued solution $r^i_f$ to \begin{equation}\label{eq:bounded_sum_simplified}
         \rho^i_f(r^i_f)^{\max\{U_f - x^i_f, x^i_f - L_f\}+1} + \rho^i_f(r^i_f)^{\min\{U_f - x^i_f, x^i_f - L_f\}+1} - (\rho^i_f + 1)r^i_f + 1 - \rho^i_f = 0.
    \end{equation}
    such that $0 < r^i_f < 1$.
    Furthermore,when used in~\eqref{eq:bound_prob}, $r^i_f$ defines a valid distribution over the support of all $\xi^i_f \in \mathbb{Z}$ that satisfy~$L_f \leq x^i_f + \xi^i_f \leq U_f$.
\end{lemma}
\begin{proof}{Proof of Lemma~\ref{lem:two_side_bound}}
    To find an~$r^i_f$ that makes~\eqref{eq:bound_prob} a probability mass function, we solve the  polynomial equation
    \begin{equation}\label{eq:bounded_sum}
    \sum_{\xi^i_f=L_f - x^i_f}^{U_f - x^i_f}\rho^i_f(r^i_f)^{|\xi^i_f|} = 1.
    \end{equation}
    Let $M = \max\{U_f - x^i_f, x^i_f - L_f\}$ and $m = \min\{U_f - x^i_f, x^i_f - L_f\}$. 
    We can regroup terms in the left hand side of the above sum to get
    \begin{equation*}
        \rho^i_f + 2\sum_{\xi^i_f = 1}^{m} \rho^i_f(r^i_f)^{\xi^i_f} + \sum_{\xi^i_f = m + 1}^{M} \rho^i_f(r^i_f)^{\xi^i_f} = 1.
    \end{equation*}
    Then, using the closed form of a finite geometric series,~\eqref{eq:bounded_sum} takes the simplified form of
    \begin{align*}
        \rho^i_f + 2\left(\frac{\rho^i_fr^i_f\left(1 - (r^i_f)^{m}\right)}{1 - r^i_f}\right) + \left(\frac{\rho^i_f(r^i_f)^{m+1}\left(1 - (r^i_f)^{M - m}\right)}{1 - r^i_f}\right) &= 1 \\
        \Leftrightarrow\;\rho^i_f - \rho^i_fr^i_f + 2\rho^i_fr^i_f - 2\rho^i_f(r^i_f)^{m+1}+ \rho^i_f(r^i_f)^{m+1} - \rho^i_f(r^i_f)^{M +1} &= 1 - r^i_f \\
        \Leftrightarrow\;\rho^i_f(r^i_f)^{M+1} + \rho^i_f(r^i_f)^{m+1} - (\rho^i_f + 1)r^i_f + 1 - \rho^i_f &= 0.
    \end{align*}
    which is exactly~\eqref{eq:bounded_sum_simplified}.
    
    We now show that the a real-valued $r^i_f \in (0,1)$ exists.
    Let~$f(r^i_f)$ be the function on the left-hand side of~\eqref{eq:bounded_sum_simplified}
    \begin{equation*}
        f(r^i_f) := \rho^i_f(r^i_f)^{M+1} + \rho^i_f(r^i_f)^{m+1} - (\rho^i_f + 1)r^i_f + 1 - \rho^i_f
    \end{equation*}
    that we wish to find real roots of. Thus,~$f(0) = 1 - \rho^i_f$ and~$f(1) = 0$. The first derivative of~$f$ is \begin{equation}\label{eq:bounded_sum_simplified_deriv}
     f'(r^i_f) = (M+1)\rho^i_f(r^i_f)^M + (m + 1)\rho^i_f(r^i_f)^m - (\rho^i_f + 1).
    \end{equation}
    Thus, $f'(1) = (M+m+1)\rho^i_f - 1 = (U_f - L_f + 1) \rho^i_f - 1$. For $\rho^i_f > \frac{1}{U_f - L_f + 1}$,  we see that $f'(1) > 0$. Thus, there must exist an $\bar{r} \in (0,1)$ such that $f(\bar{r}) < 0$. And since $f(0) > 0$, by Bolzano's Theorem, there must exist an $r^i_f \in (0,1)$ that satisfies $f(r^i_f) = 0$. \hfill \Halmos
\end{proof}
Similar to section~\ref{sec:one_sided_bound}, the hypothesis test is set up with the found~$r^i_f$ values through Lemma~\ref{lem:two_side_bound}. The hypothesis test is~\eqref{eq:bounded_test_sum}, and the tuned parameters of uncertainty set~\eqref{eq:generalUncertainty} are~$\gamma^i_f = \log \frac{1}{r^i_f}$ and~$\epsilon = - |\mathcal{I}| \log \lambda$.

Note that binary features are a special case of integer features with lower bound 0 and upper bound 1. Using the truncated symmetric geometric distribution~\eqref{eq:bound_prob} with~$\frac{1}{2} \leq \rho^i_f \leq 1$, a perturbation~$\xi^i_f$ is distributed as
\begin{equation} \label{eq:hyptest_binary}
    \mathbb{P}(\xi^i_f = \zeta^i_f) := \rho^i_f\left(\frac{1-\rho^i_f}{\rho^i_f}\right)^{|\zeta^i_f|} = \begin{cases}
        \rho^i_f & \text{if } |\zeta^i_f| = 0 \\
        1-\rho^i_f & \text{otherwise.}
    \end{cases}
\end{equation}
Thus, it follows that the tuned hyperparameters for an uncertainty set~\eqref{eq:generalUncertainty} with binary features ($L_f = 0$ and~$U_f = 1$) is~$\gamma^i_f = \log \frac{\rho^i_f}{1 - \rho^i_f}$ and~$\epsilon = - |\mathcal{I}| \log \lambda$.


\subsection{One-Sided Distribution Shifts}
\label{sec:one-sided_shifts}
In some applications, the direction of the distribution shift may be known. For instance, say that only nonnegative shifts in the values of the covariates are possible between training and deployment phases. Such a scenario can occur when, for example, there is a change in the framing of a survey question between training and deployment that reduces the sensitivity of a question, skewing the distribution of answers in the positive direction. In such settings, a model that uses set~\eqref{eq:generalUncertainty} hedges against infeasible shifts and the uncertainty set
\begin{equation}
    \Xi_+ := \left\{\boldsymbol \xi \in \mathbb{Z}_{+}^{|\mathcal{I}| \times |\mathcal{F}|}\ : \ \sum_{i \in \mathcal{I}} \sum_{f \in \mathcal{F}} \gamma^i_f \xi^i_f \leq \epsilon\right\}
    \label{eq:nonnegativeUncertainty}
\end{equation}
that only allows for nonnegative values of~$\boldsymbol \xi$ is more appropriate.
To solve~\hyperref[eq:single_setup]{$(\mathcal{P}_{\Xi_+})$}, we can use the same approach as described in section~\ref{sec:solutionmethod}, only changing the subproblem slightly by restricting the perturbations~$\boldsymbol \xi^i$ considered to be in~$\mathbb{Z}_{+}^{|\mathcal{F}|}$ for each~$i \in \sets I$ (namely, in minimization problem~\eqref{eq:cost_perturb}). A similar uncertainty set can likewise be defined to allow only nonpositive shifts or to allow a mixture of different one-sided shifts, where~\eqref{eq:single_setup} with such an uncertainty set can be solved with a similar adaptation to the subproblem.

For unbounded features, the uncertainty set calibration method in section~\ref{sec:stat_connect} can be applied in the same way as the two-sided perturbation method, as the hypothesis test is on the magnitude of perturbation regardless of direction of perturbation. For bounded features, the uncertainty set calibration method in section~\ref{sec:bounded} can also be used by setting~$L_f$ to the value of $x^i_f$.

\subsection{Distribution Shifts on Categorical Features}
\label{sec:categorical}

With some modifications, the modeling and solution approaches in this paper can be applied to handle general (not necessarily binary) categorical features.
Assuming the entire original dataset consists of categorical features, we index categorical features in the original data in the set~$\mathcal C$ and one-hot encode all features to obtain the new dataset features $\mathcal F$, wherein features indexed in the set $\mathcal F_c \subseteq \mathcal F$ are used in the one-hot encoding of categorical feature~$c$ (i.e., the sets $\{\mathcal F_c\}_{c \in \mathcal C}$ constitute a partition of $\mathcal F$). Then only one of two scenarios can materialize for each sample~$i \in \sets I$ and categorical feature~$c \in \sets C$: either the categorical feature~$c$ is not perturbed (in which case~$\xi^i_f = 0$ for all $f \in \sets F_c$) or it is, in which case the category value is changed from $f' \in \mathcal F_c$ to $\bar{f} \in \mathcal F_c \backslash\{f'\}$, i.e., $\xi^i_{f'} = -1$ for $f' \in \sets F_c$ such that~$x^i_{f'} = 1$,~$\xi^i_{\bar{f}} = 1$ for some~$\bar{f} \in \sets F_c \backslash\{f'\}$ and~$\xi^i_{{f}} = 0$ $\forall f \in \mathcal F_c \backslash\{f', \bar{f}\}$. We characterize the values of~$\bf \xi$ by
the following uncertainty set:
\begin{equation} \label{eq:cat_uncertainty}
    \begin{aligned}
        \Xi_{\text{cat}} := \{ \ & \boldsymbol{\xi} \in \{-1, 0 , 1\}^{|\sets{I}| \times |\sets{F}|} \ : \\
        & \sum_{i \in \mathcal{I}} \sum_{f \in \sets{F}} \gamma^i_f|\xi^i_f| \leq \epsilon, \\
        &0 \leq x^i_f + \xi^i_f  &\forall{i \in \sets I}, f \in \sets F_k, c \in \mathcal{C}, \\
        &\sum_{f \in \sets F_c} x^i_f + \xi^i_f = 1 &\forall{i \in \sets I}, c \in \mathcal{C} \ \}.
    \end{aligned}
\end{equation}
where the first constraint of~\eqref{eq:cat_uncertainty} is the same as the other previously mentioned uncertainty sets, which penalize any change in feature~$f \in \sets F$ with~$\gamma^i_f$ with a total budget of~$\epsilon$ and the other constraints define values of~$\bf \xi$ that either perturb or do not perturb each sample and categorical feature. The second and third constraints of~\eqref{eq:cat_uncertainty} ensure that~$\mathbf{x} + \boldsymbol{\xi}$ maintains a one-hot encoding of each categorical feature.

When solving problem~\eqref{eq:cost_perturb} in algorithm~\ref{alg:subproblem} for each~$i \in \sets I$, the perturbations that lead down each path must satisfy the constraints on~$\boldsymbol{\xi}$ in uncertainty set~\eqref{eq:cat_uncertainty} to ensure that the perturbation on each categorical feature corresponds to a change of value in the one-hot encoding.
In other words, for each categorical feature~$c \in \sets C$ that is perturbed in sample~$i$, we must have~$\xi^i_{f'} = -1$ and~$\xi^i_{\bar f} = 1$ for two distinct~$f',\bar{f} \in \sets F_c$. So by the left-hand side of the first constraint of~\eqref{eq:cat_uncertainty}, perturbing one categorical feature incurs a cost of perturbation of~$\gamma^i_{f'} + \gamma^i_{\bar{f}}$. All else in our solution method remains the same.

To calibrate the uncertain parameters of~\eqref{eq:cat_uncertainty}, we again note that due to the second and third constraints in~\eqref{eq:cat_uncertainty}, any perturbation of one feature~$f' \in \sets F_c$ will cause an equal and opposite perturbation in some other feature~$\bar{f} \in \sets F_c \backslash \{f'\}$. 
We assume that the probability of no perturbation in categorical feature $c \in \sets C$ of sample~$i$ is~$\frac{1}{|\sets F_c|} \leq \rho^i_c \leq 1$, and that the realizations of all other values are all equally likely. Similar to the bounded integer case (see section~\ref{sec:two_sided_bounds}), we place the lower bound of~$\frac{1}{|\sets F_c|}$ on~$\rho^i_c$ -- this allows for maximal uncertainty on the value of~$c$. Then, the probability of the perturbation of categorical feature $c \in \sets C$ encoded by the collection of perturbations~$\{\xi^i_f\}_{f \in \sets F_c}$ on the one-hot encoding of~$c$, is 
\begin{equation} \label{eq:cat_prob}
        \mathbb{P}(\{\xi^i_f\}_{f \in \sets F_c} = \{\zeta^i_f\}_{f \in \sets F_c}) :=
        \mathbb{I}\left[\sum_{f \in \sets F_c} |\zeta^i_f| = 0\right]\rho^i_c + \mathbb{I}\left[\sum_{f \in \sets F_c} |\zeta^i_f| = 2\right]\left(\frac{1 - \rho^i_c}{|\Phi_k| - 1}\right).
\end{equation}
As done before, we set up our hypothesis test as a likelihood ratio test with threshold~$\lambda^{|\mathcal{I}|}$:
\begin{equation}
    \frac{\prod_{i \in \sets{I}} \prod_{c \in \sets{C}} \mathbb{P}(\{\xi^i_f\}_{f \in \sets F_c})}{\prod_{i \in \sets{I}} \prod_{c \in \sets{C}} \mathbb{P}(\{0\}_{f \in \sets F_c})} \geq \lambda^{|\mathcal{I}|}.
\end{equation}
Plugging in (29) into (30) yields
\begin{equation}~\label{eq:cat_test_unsimplified}
   \sum_{i \in \sets{I}} \sum_{c \in \sets{C}} -\log\left( \mathbb{I}\left[\sum_{f \in \sets F_c} |\xi^i_f| = 0\right] + \mathbb{I}\left[\sum_{f \in \sets F_c} |\xi^i_f| = 2\right]\left(\frac{1 - \rho^i_c}{\rho^i_c(|\sets F_c| - 1)}\right)\right) \leq -|\mathcal{I}| \log \lambda.
\end{equation}
Thus, if $\sum_{f \in \sets F_c} |\xi^i_f| = 0$, i.e., no perturbation occurs, no cost is incurred; if $\sum_{f \in \sets F_c} |\xi^i_f| = 2$, i.e., a perturbation occurs, then a cost of $-\log\left(\frac{1 - \rho^i_c}{\rho^i_c(|\sets F_c| - 1)}\right)$ incurred. We can represent~\eqref{eq:cat_test_unsimplified} more compactly as
\begin{equation}~\label{eq:cat_test_simple}
   \sum_{i \in \sets{I}} \sum_{f \in \mathcal{F}}  \frac{1}{2}\log\left(\frac{\rho^i_c(|\{\sets F_c\}| - 1)}{1 - \rho^i_c}\right)|\xi^i_f| \leq -|\mathcal{I}| \log \lambda.
\end{equation}
It follows from the above that the tuned hyperparameters in the uncertainty set~\eqref{eq:cat_uncertainty} are~$\gamma^i_f = \frac{1}{2}\log\left(\frac{\rho^i_c(|\{\sets F_c\}| - 1)}{1 - \rho^i_c}\right)$ and~$\epsilon = - |\mathcal{I}| \log \lambda$. 

We note that if~$c$ is binary and one-hot encoded into two features following the process described in this section, the tuned uncertainty set parameter values are equivalent to that described in section~\ref{sec:two_sided_bounds}.

\subsection{Mixed Datasets and Distribution Shifts}

Often, datasets have a mixture of unbounded integer, bounded integer, binary, and categorical data, and known directions of distribution shifts vary across features. In such scenarios, we can adapt the models and calibration methods presented in section \ref{sec:uncertainty_set} and \ref{sec:stat_connect}, as well as the Electronic Companion sections~\ref{sec:bounded},~\ref{sec:one-sided_shifts}, and~\ref{sec:categorical} to create uncertainty sets that capture this information.

Consider the following uncertainty set
\begin{equation}
    \Xi_{\rm{mixed}} := \left\{\boldsymbol \xi \in \mathcal{M} : \ \sum_{i \in \mathcal{I}} \sum_{f \in \mathcal{F}} \gamma^i_f \vert \xi^i_f \vert \leq \epsilon \right\}
\end{equation}
where~$\mathcal{M} \subseteq \mathbb{Z}^{|\sets{I}| \times |\sets{F}|}$ can be use to place restrictions on the kinds of shifts allowed for each datapoint (e.g., to capture one-sided perturbations, bounded shifts, or categorical features). 
Letting~$\rho^i_f$ denote the probability of certainty in the nominal value of the data, the tuned hyperparameters are~$\epsilon = -|\mathcal{I}| \log \lambda$ and
\begin{equation}
    \gamma^i_f := \begin{cases}
        \log\left(\frac{1}{1-\rho^i_f}\right) & \text{for} \ f \ \text{an unbounded integer} \\
        \log\left(\frac{1}{r^i_f}\right) & \text{for} \ f \ \text{a bounded integer}\\
        \log\left(\frac{\rho^i_f}{1-\rho^i_f}\right) & \text{for} \ f \ \text{binary}\\
        \frac{1}{2}\log\left(\frac{\rho^i_f(|\{\Phi_k\ : \ f \in \Phi_k\}| - 1)}{1 - \rho^i_f}\right) & \text{for} \ f \ \text{part of a categorical feature,}
    \end{cases}
\end{equation}
where $0 < r^i_f < 1$ is found by Lemma~\ref{lem:one_side_bound} for one-sided bounds or Lemma~\ref{lem:two_side_bound} for two-sided bounds (see Electronic Companion~\ref{sec:bounded}). The solution method described in section~\ref{sec:solutionmethod} remains the same, only needing to adapt the subproblem algorithm~\ref{alg:subproblem} to only allow perturbations from the set~$\mathcal{M}$.

\section{Numerical Comparison to Alternative Robust Method}
\label{appendix:experiment_bertsimas}
We empirically analyze the performance of our method against the method of~\cite{Bertsimas2019RobustClassification}, see section~\ref{sec:comparisons} for comparisons between each model. For simplicity, we will refer to our method as \texttt{\rct}  and the method of~\cite{Bertsimas2019RobustClassification} as~\texttt{\brct} in this section. 

\subsection{Setup}
For fair comparison to the method of~\cite{Bertsimas2019RobustClassification}, we only evaluate \texttt{\brct} on instances corresponding to datasets in Table~\ref{tab:datasets} with only binary features (i.e., \texttt{house-votes-84} and \texttt{spect}). 
For datasets with non-binary features, \texttt{\brct} cannot tackle problems with the uncertainty sets we consider in this paper.

To adapt our uncertainty set to a comparable uncertainty set in the \texttt{\brct} formulation, we first transformed the uncertainty set used in the above experiment (i.e. the uncertainty set~\eqref{eq:generalUncertainty} tuned using~\eqref{eq:hyptest_binary}) into the uncertainty set~\eqref{eq:bertsimas_uncertainty} for $p = \infty$. For each instance of our original problem, let~$\rho$ be the average probability of certainty used to generate the~$\rho^i_f$ values in the experiments on our methods.
Then,~\eqref{eq:hyptest_binary} becomes for each $i \in \mathcal{I}$
\[
    \sum_{f \in \sets F} |\xi^i_f| \leq -\frac{\log \lambda}{\log\left(\frac{\rho}{1-\rho}\right)}.
\]
This matches the form of~\eqref{eq:bertsimas_uncertainty} for $p = 1$, and therefore can be used to define the parameters of~\eqref{eq:bertsimas_uncertainty} and have a more similar comparison to our methodology.
Additionally, we apply no regularization term for all experiments and set a time limit of 7200 seconds, which is the same as in our formulation.

\subsection{Accuracy Under Distribution Shifts}
\label{sec:accuracy_bertsimas}
\begin{figure}[t!]
\centering
\begin{subfigure}[b]{0.49\textwidth}
\centering
\includegraphics[width=\textwidth]{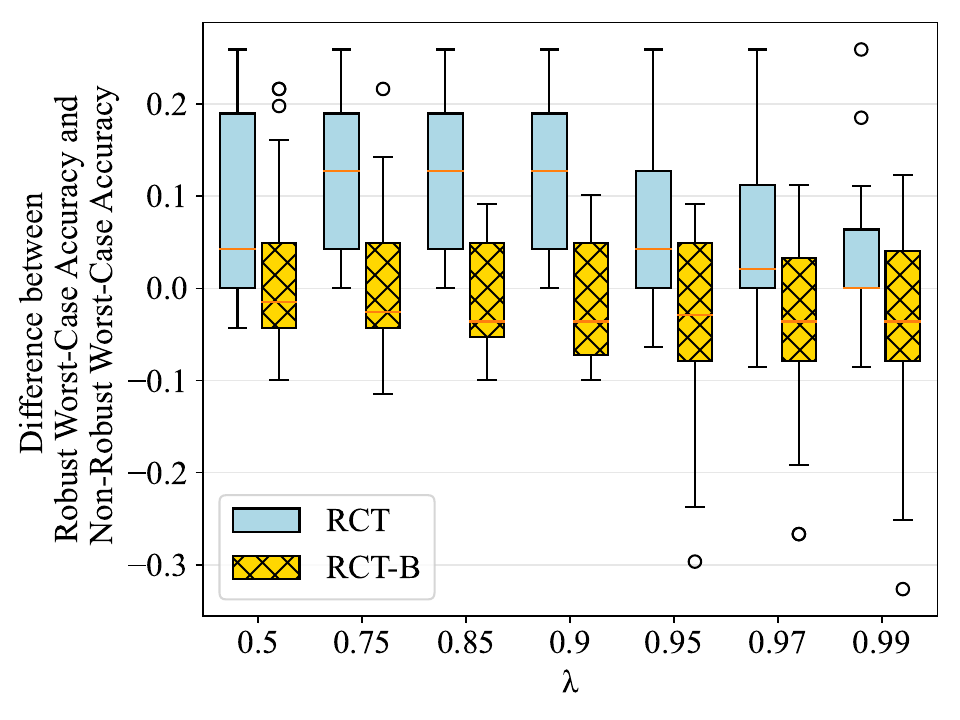}
\caption{}
\end{subfigure}
\begin{subfigure}[b]{0.49\textwidth}
\centering
\includegraphics[width=\textwidth]{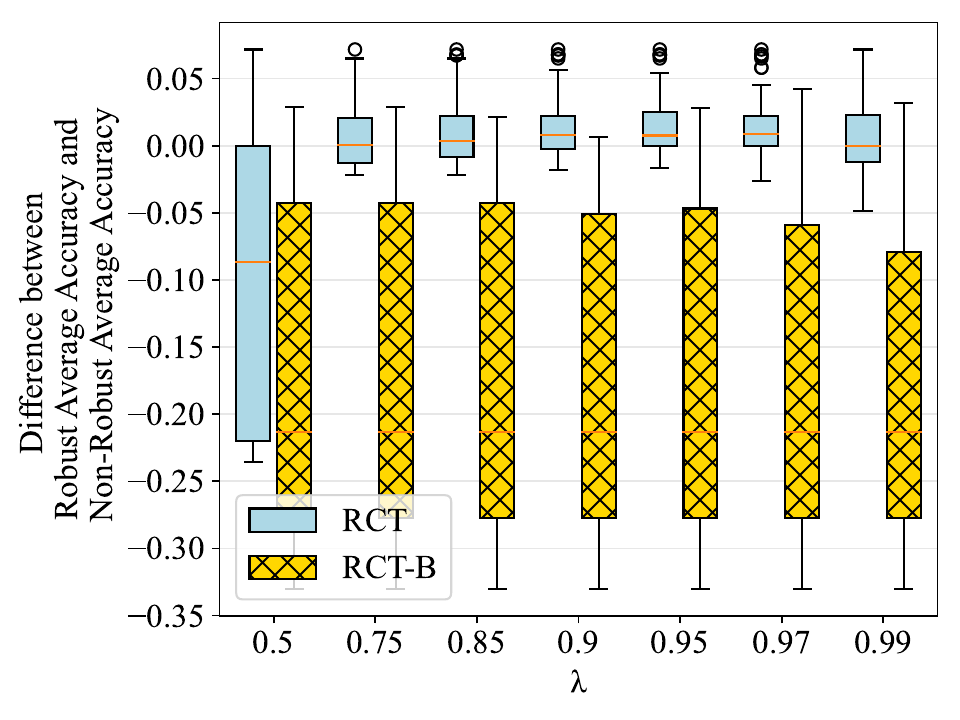}
\caption{}
\end{subfigure}
\caption{Companion figure to section~\ref{sec:accuracy_bertsimas}. 
Distribution across problem instances of the gain in (a) worst-case accuracy and (b) average accuracy from using \texttt{\rct} versus \texttt{\brct} in dependence of~$\lambda$ on binary datasets. The parameter~$\lambda$ varies between 0.5 (more robust) and 0.99 (less robust).}
\label{fig:boxacc_bertsimas}
\end{figure}

We analyze the worst-case and average-case accuracies on the same 5000 perturbed datasets for each instance of \texttt{\rct} and its analogous instance of \texttt{\brct}. The results for the worst-case and average case on expected perturbations are shown in Figure~\ref{fig:boxacc_bertsimas}. We see that on binary datasets, there is on average a decrease in worst-case performance in the \texttt{\brct} method across all values of~$\lambda$ in comparison to the non-robust tree, whereas \texttt{\rct} has on average a favorable gain in worst-case performance on the same instances. Specifically, \texttt{\rct} returns a higher gain in worst-case performance than that of \texttt{\brct} by up to about 15\% on average in the worst-case. In addition, the average-case performance of \texttt{\brct} suffers in comparison to the non-robust tree on average, whereas \texttt{\rct}  on average remains on par with the average-case performance of a non-robust tree in the instances where~$\lambda$ is 0.75 or above. These results match the conclusions from proposition~\ref{prop:subset}, where \texttt{\brct} is more conservative than \texttt{\rct}, potentially causing a bad performance in testing.

\subsection{Price of Robustness}
\label{sec:price_robustness_bertsimas}
\begin{figure}[t!]
\centering
\begin{subfigure}[b]{0.49\textwidth}
\centering
\includegraphics[width=\textwidth]{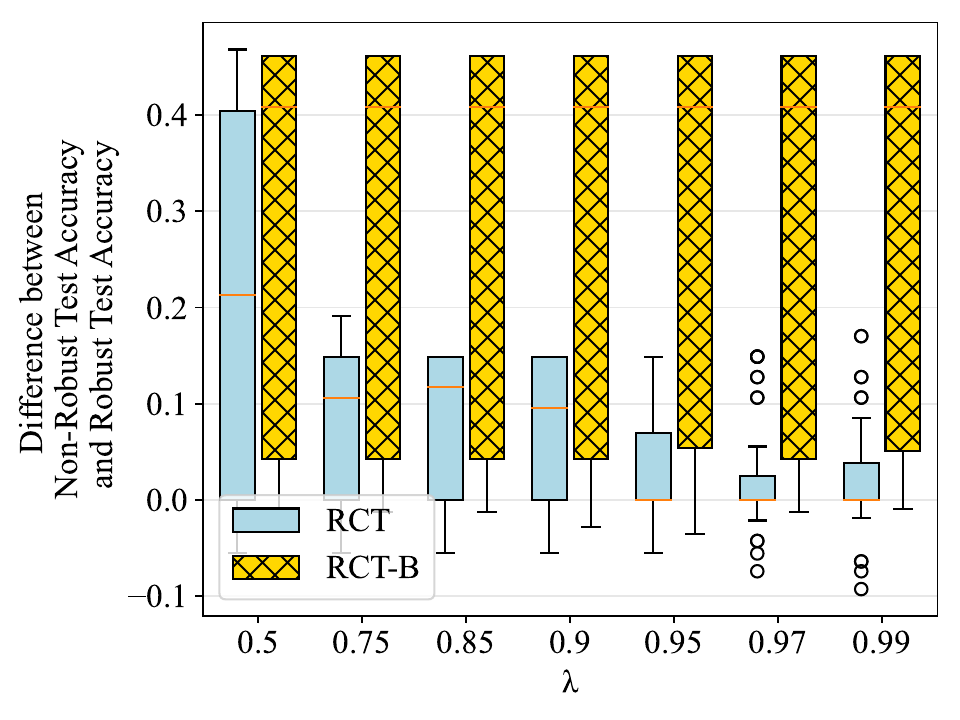}
\caption{}
\label{fig:boxprice_compare}
\end{subfigure}
\begin{subfigure}[b]{0.49\textwidth}
\centering
\includegraphics[width=\textwidth]{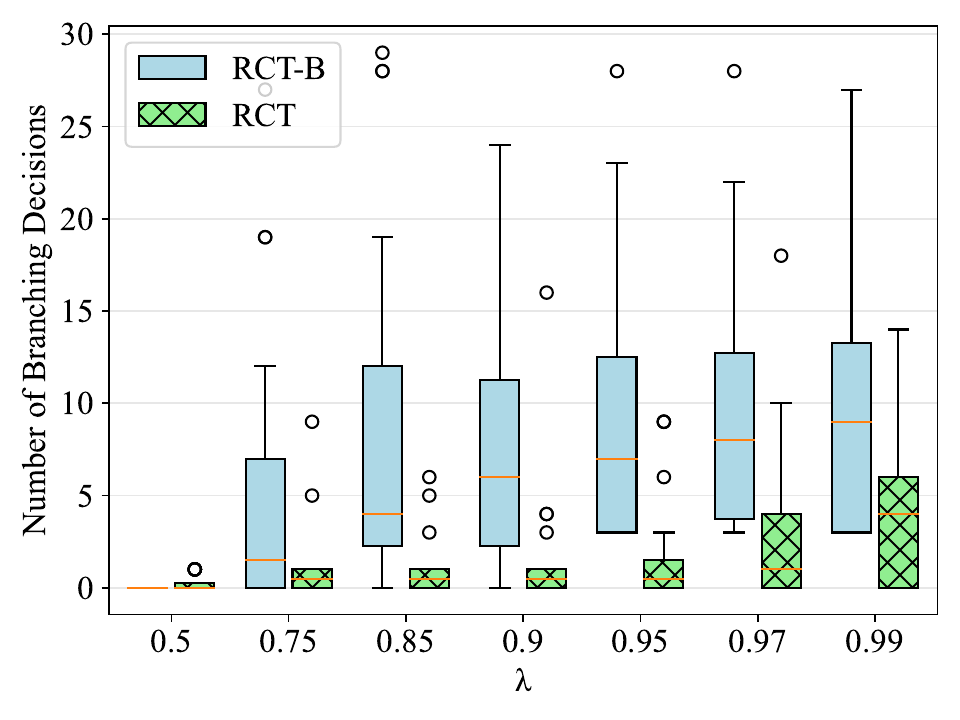}
\caption{}
\label{fig:num_branching_comparison}
\end{subfigure}
\caption{(a) Companion figure to section~\ref{sec:price_robustness_bertsimas}. Distribution across problem instances of the loss in nominal test accuracy from using \texttt{\rct} versus \texttt{\brct} in dependence of~$\lambda$ on binary datasets. The parameter~$\lambda$ varies between 0.5 (more robust) and 0.99 (less robust). (b) Companion figure to section~\ref{sec:sparsity_bertsimas}. Distribution of the number of branching nodes across all instances with the same budget of uncertainty for both \texttt{\rct} and \texttt{\brct}.}
\end{figure}

For both \texttt{\rct} and \texttt{\brct}, we measure the decrease in accuracy from using a robust tree versus a non-robust tree under the case of no distribution shift in the test set, i.e., 
the price of robustness \citep{bertsimas2004price}, and summarize this metric in Figure~\ref{fig:boxprice_compare}.
From the figure, we observe that for each value of~$\lambda$, \texttt{\rct} has an average a price of robustness less than $0.25$. In contrast, \texttt{\brct} has an average price of robustness of about $0.4$ for all~$\lambda$.

\subsection{The Effect of Robustness on Sparsity}
\label{sec:sparsity_bertsimas}
For each value of~$\lambda$ in Figure~\ref{fig:num_branching_comparison}, we compare the number of branching nodes of \texttt{\rct} against the method of \texttt{\brct}, aggregating across each comparable instance between the two methods. We first note that the number of branching nodes of \texttt{\brct} remains relatively consistent for values of~$\lambda$ above 0.75, up until the budget of uncertainty becomes too large (i.e.,~$\lambda = 0.5$) where many instances yield a tree with no branching nodes. As previously mentioned, the method of~\cite{Bertsimas2019RobustClassification} constrains every sample to have the same predictions across all realizations of the covariates. So with a large enough uncertainty set, all samples are given the majority label regardless of the covariates, which is the same as yielding a tree with no branching nodes.

In comparison, we see that \texttt{\rct} does not have the same behavior as \texttt{\brct}, as the number of branching nodes drops smoothly from a smaller uncertainty set to a larger uncertainty set. The tree structure changes through variations in the parameters of uncertainty more consistently in \texttt{\rct} in comparison to \texttt{\brct}, suggesting that the trees generated by \texttt{\rct} are closely tailored to the specific distribution shifts it hedges against.

\subsection{Computational Times}
\label{sec:times_bertsimas}
\begin{figure}[t]
\centering
\includegraphics[width=0.49\textwidth]{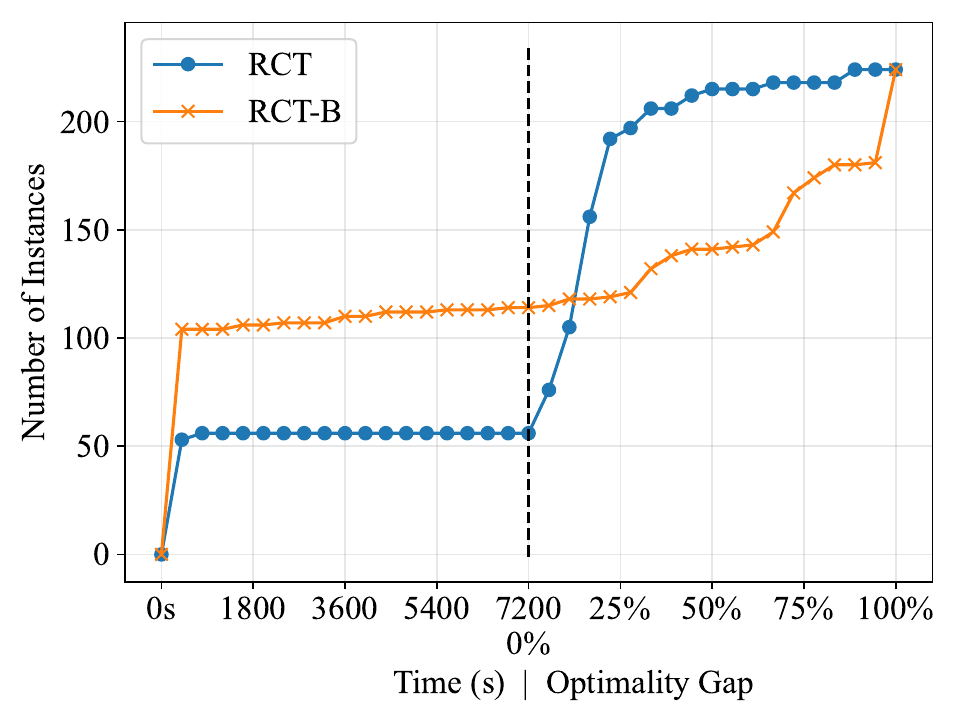}
\caption{Companion figure to section~\ref{sec:times_bertsimas}. Number of instances solved \revision{to optimality on the `Time' axis and number of instances with optimality gap no larger than each given value at the time limit on the `Optimality Gap' axis for} \texttt{\rct} and \texttt{\brct}.}
\label{fig:runtime_comparison}
\end{figure}

Figure~\ref{fig:runtime_comparison} shows the computation times and optimality gaps of all instances of \texttt{\brct} in comparison to the corresponding instances of \texttt{\rct}. We see that about two times more instances of \texttt{\brct} can be solved within the time limit in comparison to \texttt{\rct}. However, for instances that could not be solved to optimality in the time limit, the optimality gap overall is much smaller in instances of \texttt{\rct} than instances of \texttt{\brct}. There are several reasons for this. In \texttt{\brct}, instances that resulted in no branching nodes were solved very quickly, as the only feasible trees were zero-depth ones. For other instances of \texttt{\brct} where the uncertainty set did allow for several feasible solutions, the convergence to optimality is noticeably less than that \texttt{\rct}, where nearly all instances of \texttt{\rct} are within fifty percent of the optimality gap. This suggests that in \texttt{\brct} instances where a nonzero-depth tree is the optimal tree, there is a greater computational costs of learning trees with \texttt{\brct} than with \texttt{\rct}.

\end{document}